\documentclass{article}

\usepackage[preprint]{neurips_2026}

\usepackage[utf8]{inputenc} 
\usepackage[T1]{fontenc}    
\usepackage{hyperref}       
\usepackage{url}            
\usepackage{booktabs}       
\usepackage{amsfonts}       
\usepackage{nicefrac}       
\usepackage{microtype}      
\usepackage{xcolor}         
\usepackage[utf8]{inputenc} 
\usepackage[T1]{fontenc}    
\usepackage{hyperref}       
\usepackage{url}            
\usepackage{booktabs}       
\usepackage{amsfonts}       
\usepackage{nicefrac}       
\usepackage{booktabs}
\usepackage[table]{xcolor}
\usepackage{graphicx}
\usepackage{microtype}      
\usepackage{xcolor}         
\usepackage{graphicx}
\usepackage{subcaption}

\usepackage{amsmath,amsfonts,bm}

\DeclareMathOperator{\NLR}{NLR}

\def\eqref#1{equation~\ref{#1}}

\def\1{\bm{1}}

\DeclareMathAlphabet{\mathsfit}{\encodingdefault}{\sfdefault}{m}{sl}
\SetMathAlphabet{\mathsfit}{bold}{\encodingdefault}{\sfdefault}{bx}{n}

\usepackage{amsmath,amssymb,amsthm}

\usepackage{hyperref}
\usepackage{pifont}
\usepackage{url}
\usepackage{xspace}
\usepackage{amsmath,amssymb,amsthm}
\usepackage{mathtools}
\usepackage{enumitem}
\usepackage{comment}
\usepackage{stfloats}
\usepackage{natbib}
\usepackage{xspace}
\usepackage{color, colortbl}
\usepackage{algorithm}
\usepackage{algpseudocode}
\usepackage{multirow}

\ifx\figurename\undefined \def\figurename{Figure}\fi
\renewcommand{\figurename}{Fig.}
\renewcommand{\paragraph}[1]{\textbf{#1} }

\newcommand{\cmark}{\ding{51}}%
\newcommand{\xmark}{\ding{55}}%
\newcommand{\Sect}[1]{Sec.~\ref{#1}}
\newcommand{\Fig}[1]{Fig.~\ref{#1}}
\newcommand{\Tbl}[1]{Tbl.~\ref{#1}}
\newcommand{\Eqn}[1]{Eqn.~\ref{#1}}
\newcommand{\Apx}[1]{Apdx.~\ref{#1}}

\newcommand{\method}{DynaDiag\xspace}

\newcommand{\MethodName}{\textsc{ShuffleSparse}\xspace}

\newcommand{\no}[1]{#1}
\renewcommand{\no}[1]{}
\newcommand{\RNum}[1]{\uppercase\expandafter{\romannumeral #1\relax}}

\title{\MethodName: Learned Shuffles for Structured Sparse Networks}

\author{%
  Abhishek Tyagi\\
  Department of Computer Science\\
  University of Rochester\\
  Rochester, NY 14620 \\
  \texttt{atyagi2@cs.rochester.edu} \\
   \And
  Arjun Iyer \\
  The Institute of Optics\\ 
  University of Rochester \\
  Rochester, NY 14620 \\
   \texttt{aiyer2@ur.rochester.edu} \\
  \AND
  Liam Young\\
  The Institute of Optics\\ 
  University of Rochester \\
  Rochester, NY 14620 \\
   \texttt{lyoung12@ur.rochester.edu}\\
   \And
   William H. Renninger\\
   The Institute of Optics\\ 
  University of Rochester \\
  Rochester, NY 14620 \\
   \texttt{wrenning@ur.rochester.edu} \\
   \And
   Christopher Kanan\\
   Department of Computer Science\\ 
  University of Rochester \\
  Rochester, NY 14620 \\
   \texttt{ckanan@cs.rochester.edu} \\
   \And
   Yuhao Zhu\\
   Department of Computer Science\\ 
  University of Rochester \\
  Rochester, NY 14620 \\
   \texttt{yzhu@rochester.edu} \\
}

\begin{document}

\maketitle

\begin{abstract}

Structured weight sparsity accelerates training and inference on modern GPUs, but it trails unstructured dynamic sparse training (DST) in accuracy especially at extreme sparsity. We pinpoint the reason for this difference in performance to a lack of expressivity: a dense layer can implement any pattern of non-zero weights, whereas structured patterns are restricted to only a small set of weight configurations. We introduce \MethodName, a single permutation primitive that applies uniformly across DST-from-scratch and one-shot pruning, and across N:M, block etc. We close most of this gap by learning a single permutation matrix jointly with the structured weight matrix. When used on three different types of structures (block, N:M, and diagonal), \MethodName is able to reduce the structured-vs-unstructured accuracy gap on ViT-B/16 (ImageNet-1K) and GPT-2 (WikiText-103) at 90–95\% sparsity, while adding minimal inference overhead ($\leq 8.7$\% inference overhead) and preserving any training acceleration the host structure provides. The same permutation formulation transfers to one-shot 2:4 pruning of pretrained LLMs, where it improves zero-shot accuracy by 4.6 points on LLaMA-2 7B. Together, these results establish learned permutations as a general tool for recovering unstructured-level accuracy from structured sparsity patterns.
\end{abstract}

\section{Introduction}
\label{sec:intro}


Deep neural networks (DNNs) have been expanding over the years, with their capabilities on complex tasks reaching human-level performance~\citep{labryth, chess}. At the same time, the cost of training and inference for such large DNNs has increased drastically~\citep{cottier2023trends}. A principled way to reduce this cost, while retaining high accuracy, is to exploit sparsity by removing unnecessary weights via pruning~\citep{molchanov2016pruning, tanaka2020pruning} or training directly in the sparse regime~\citep{jaiswal2022training,zhang2023universal}, often achieving similar algorithmic performance as dense models under comparable budgets~\citep{frankle2018lottery,blalock2020state,mostafa2019parameter}. In both settings, turning theoretical FLOP savings into wall-clock speedup on modern accelerators requires the weights to follow a \emph{structured} pattern like $N{:}M$, block, or diagonal that matches vendor-optimized kernels~\citep{mishra2021accelerating,liu2019improving}.

Structured sparsity aligns well with vendor-optimized kernels, and is competitive with unstructured methods at low to moderate sparsity. But it commonly underperforms its unstructured counterpart especially at \emph{extreme} sparsities (90\% and above) because the rigid patterns reduce the expressivity of the network. This gap shows up consistently in both regimes: structured dynamic sparsity training (DST) methods and one-shot pruning. 

One solution is to learn a \emph{permutation} of the input dimensions, in which case a structured pattern in the permuted basis corresponds to maps that are totally unstructured in the original basis. The idea of this has been studied in parts. \emph{AutoShuffleNet}~\citep{lyu2020autoshufflenet} learns channel permutations in \emph{dense} ShuffleNet architectures. NVIDIA's permutation search~\citep{pool2021channel} finds \emph{static} permutations for fixed 2:4 weights via discrete optimization. \emph{PermLLM}~\citep{zou2025permllm} learns block-wise channel permutations during \emph{pruning} of pretrained LLMs. Each method is tied to a specific setting and a specific sparsity pattern; no prior work has shown that a single permutation primitive works across \emph{both} training-from-scratch and pruning, and across the canonical structured patterns.


\textbf{A single learned permutation per layer is the unifying primitive for structured sparsity, generalizing across both training-from-scratch and pruning of pretrained networks.} Our method, \MethodName, learns one permutation per sparsified layer jointly with its structured weight matrix via a soft relaxation that hardens by the end of training, and the same formulation plugs into N:M, block, and diagonal patterns without method-specific changes. Empirically, \MethodName shrinks the gap to unstructured baselines on ViT-B/16 and GPT-2 in DST, and outperforms the strongest non-permuted 2:4 pruning baseline by 4.6 points on LLaMA-2 7B. In this work, we make the following contributions:
\begin{enumerate}\setlength\itemsep{2pt}
    \item We introduce \MethodName, a single layer formulation that 
    applies to both dynamic sparse training from scratch and pruning of pretrained LLMs, and works uniformly across structured sparse patterns.
    \item Across four structured DST methods (SRigL, PixelatedBFly, DSB, 
    DynaDiag) on ViT-B/16 (ImageNet-1K) and GPT-2 Small/Medium 
    (WikiText-103), \MethodName consistently shrinks the gap to 
    unstructured baselines. The strongest pairing, reaches a 0.97\% average accuracy gap on ViT-B/16 (down from 1.92\%) and 1.50\% / 2.63\% perplexity gap on 
    GPT-2 Small/Medium (down from 2.27\% / 3.75\%).
    \item Applied to frozen pretrained LLMs at 2:4 
    sparsity, the same permutation formulation improves over the 
    strongest non-permuted baseline by \textbf{+4.62} average points on LLaMA-2 7B and \textbf{+2.95} on Qwen-2.5 3B.
\end{enumerate}

\section{Related Work}
\label{sec:related}

\textbf{Sparse training.}
Pruning removes small-magnitude or low-salience weights from a pre-trained dense model and then fine-tunes it, still paying the full dense pre-training cost \citep{molchanov2017variational, cai2022structured, lin2019towards, lu2024alphapruning}.
The Lottery Ticket Hypothesis~\citep{frankle2018lottery} motivates training sparse models from scratch, which falls into \emph{static} regimes that fix a mask at initialization (e.g., Pixelated Butterfly~\citep{dao2021pixelated}) and \emph{dynamic} regimes that update connectivity via prune-and-grow rules.
Works such as SET prunes by magnitude and regrows randomly \citep{mocanu2018scalable}; MEST mixes magnitude and gradient signals \citep{yuan2021mest}; RigL prunes by magnitude and regrows by gradient on missing connections~\citep{evci2020rigging}; recent work explores topology-driven, gradient-free growth and N:M-style constraints for scalability \citep{zhang2024epitopological,zhang2023universal}.

\textbf{Permutation learning and channel mixing.}
Beyond fixed channel shuffles, \emph{AutoShuffleNet}~\citep{lyu2020autoshufflenet} learns channel permutations \emph{during training} by optimizing a Lipschitz-continuous penalty that drives a stochastic matrix toward a permutation.
The \emph{Kaleidoscope}~\citep{dao2020kaleidoscope} (K-matrix) framework provides a differentiable, expressive parameterization that includes permutations, and it has been used to \emph{learn latent permutations} in permuted-image tasks within end-to-end training.
Pool et al.~\citep{pool2021channel} use offline permutations to prune networks to a fixed N:M sparsity pattern.
Most closely related, \emph{PermLLM}~\citep{zou2025permllm} learns block-wise channel permutations via Sinkhorn normalization, but is a \emph{post-training} method restricted to N:M pruning of pre-trained LLMs and optimizes permutations alone on frozen weights.
In contrast, we learn permutations \emph{jointly with sparse weights from scratch}, and our method is pattern-agnostic, accommodating block, N:M, and diagonal structures within a single framework, leading to more hardware-efficient sparse networks.



\section{\MethodName: Learning Permutations for Structured Sparse Networks}
\label{sec:method}

\begin{figure*}[t]
	\centering
	\includegraphics[width=\textwidth]{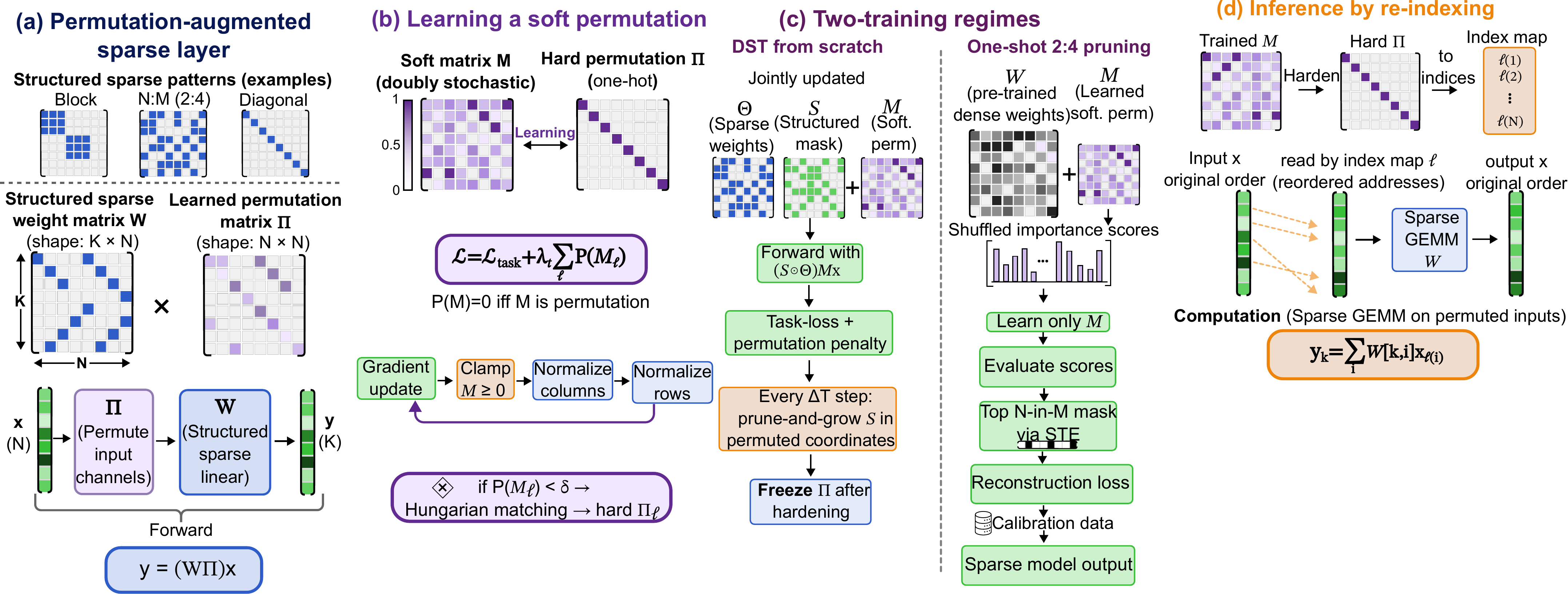}
    \caption{Overview of \MethodName. a) Each layer is composed of a structured sparse pattern with a learned permutation on the input channels. b)The discrete permutation is relaxed to a doubly-stochastic matrix and optimized jointly with the task loss under a penalty. Once the penalty drops below a threshold, the soft matrix is hardened to a one-hot permutation via Hungarian matching and frozen. c) \emph{Left}: In DST, the sparse mask and permutation are co-learned, with prune-and-grow steps applied in the permuted coordinates. \emph{Right:} in one-shot 2:4 pruning of pre-trained LLMs, the dense weights are frozen and only the permutation is trained against a calibration-set reconstruction loss. }
\label{fig:overview}
\end{figure*}

In \Sect{sec:method:layer}, we elaborate on \MethodName formulation and our intuition about why permutations can be useful. We then show in \Sect{sec:method:dst} and \Sect{sec:method:prune}, that \MethodName can be applied to two different settings of sparse models landscape: dynamic sparse training (DST) from scratch, and N:M pruning of pretrained LLMs. Lastly, \Sect{sec:method:inf} explains how permutations are executed at inference and why it incurs some overhead. \Fig{fig:overview} summarizes our method \MethodName.

\subsection{Layer Formulation}
\label{sec:method:layer}

Our aim is to co-learn permutations which can be appended to any structured-sparse layer to increase its expressivity. For a sparsified layer with weight $W\in \mathbb{R}^{R \times C}$ and input $x$, the layer is parameterized as
\begin{equation}
    y \;=\; (W\Pi)\,x \;=\; W'\,x, \qquad \Pi \in \mathcal{P}_d,
    \label{eq:form}
\end{equation}
where $W$ has a fixed structured-sparse pattern $\mathcal{S}$ (block, $N{:}M$, diagonal etc.) and $\Pi \in \mathcal{P}_d$ with $d = \min\{R,C\}$ is a learned column permutation\footnote{Row permutations ($y = \Pi W x$) based networks have the same parameter count and similar optimization behavior. We found no statistically significant accuracy difference between the two (Sec.~\ref{sec:results:abl}) and use column permutations because they admit a cheaper inference path (Sec.~\ref{sec:method:inf})}. In \Apx{sec:analysis} we analyze structured-sparse ReLU networks through the number of linear regions (NLR) framework~\citep{montufar2014number}. Without per-layer mixing, axis-aligned sparsity caps the effective dimension at the structure's rank cap $r_{\text{struct}}$, stalling NLR's depth-multiplicative growth. A full-rank permutation per layer injects $r_{\text{struct}}$ fresh directions and restores this growth after $\lceil d_0 / r_{\text{struct}} \rceil$ layers, motivating \Eqn{eq:form}.

Despite its simplicity, this parameterization has two properties that make it useful across the structured-sparse families we consider. \emph{(i) It works with any structure.} Nothing in the formulation depends on what $\mathcal{S}$ is. We demonstrate this empirically by applying \MethodName{} without method-specific changes on top of block-sparse (PixelatedBFly, DSB), $N{:}M$ (Wanda, SparseGPT), diagonal (DynaDiag), and per-row-fan-in (SRigL) baselines (Sec.~\ref{sec:results}). \emph{(ii) The hardware speedup of structured sparsity is preserved.} One of the reasons structured sparsity is interesting is that specialized GPU kernels can compute the forward GEMM faster than dense. The same speedup is available in the backward pass only if the transposed weight $W^\top$ also has a structure the kernel recognizes. Since $(W\Pi)^\top = \Pi^\top W^\top$, the transposed weight is the training method's $W^\top$ with a row permutation applied:the same structure with a different ordering. So when the method's structure
is transposable (like $N{:}M$ pattern of \citep{hubara2021accelerated}), the backward pass can also be accelerated, with or without \MethodName{}. For methods whose structure is not transposable, the
backward pass remains dense; \MethodName{} does not change this behavior in either direction.

Realizing these properties requires learning $\Pi$ end-to-end, but permutation
matrices are discrete and prevent direct gradient-based optimization.

\paragraph{Differentiable permutation estimator.} To learn the permutations, we adopt the relaxation of \citep{lyu2020autoshufflenet}: rather than optimizing $\Pi$ directly, we learn a soft matrix $M \in \mathbb{R}^{N \times N}$ and decode a hard permutation at the end of permutation learning. To encourage $M$ toward such a one-hot configuration, \citep{lyu2020autoshufflenet}
add to the task loss ($\mathcal{L}_{\mathrm{task}}$) the row- and column-wise $\ell_1{-}\ell_2$ penalty $P(M)$;

\begin{equation}
\label{eq:permloss}
    P(M) \;=\;
    \sum_{i=1}^N \big(\|M_{i:}\|_1 - \|M_{i:}\|_2\big)
    \;+\;
    \sum_{j=1}^N \big(\|M_{:j}\|_1 - \|M_{:j}\|_2\big).
\end{equation}
where $M_{i:}$ and $M_{:j}$ denote the $i$-th row and $j$-th column. For any nonnegative vector $v$, the difference $\|v\|_1 - \|v\|_2$ is zero iff $v$ is one-hot, so $P(M) = 0$ iff every row and column of $M$ is one-hot. This, combined with the doubly-stochastic constraints, means $M$ is a permutation matrix~\citep{lyu2020autoshufflenet}. Adding $P(M)$ to the task loss therefore
pushes $M$ toward becoming a valid hard permutation:
\begin{equation}
\label{eq:totalloss}
    \mathcal{L} \;=\; \mathcal{L}_{\mathrm{task}}
    \;+\; \lambda_t \sum_\ell P(M_\ell),
\end{equation}
where $M_\ell$ is the per-layer soft permutation associated with the $\ell$-th sparsified layer's $\Pi_\ell$, and $\lambda_t$ is a penalty weight ramped linearly from $0$ to $\lambda_{\max}$ over the first $T_{\mathrm{perm}}$ training steps so that $M_\ell$ first explores freely and then sharpens.



To show generalizability of the formulation above, we use it in two settings of sparsifying a network: dynamic sparse training, where the mask evolves jointly with the permutation (\Sect{sec:method:dst}), and 2:4 pruning of pretrained LLMs, where the mask is fixed by a one-shot pruning method before the permutation is learned (\Sect{sec:method:prune}). The two settings share the layer formulation, the loss in Eq.~\ref{eq:totalloss}, and hardening of the soft permutation; they differ only in how the mask is produced and in the schedule that interleaves mask updates with permutation learning.

\subsection{Dynamic Sparse Training (DST) With \MethodName}
\label{sec:method:dst}
In DST, the sparse weight $W_\ell = S_\ell \odot \Theta_\ell$ for a layer $l$, the structured mask $S_\ell$, and the permutation $M_\ell$ (we represent soft-permutation with $M$ and hard permutation with $\Pi$) are all learned jointly under a fixed global sparsity budget. The mask $S_\ell$ satisfies the chosen structure (block, $N{:}M$, diagonal, banded), and the permutation parameters are not counted toward the sparsity budget.

The permuted sparse layer computes
\begin{equation}
\label{eq:dst-forward}
    y_\ell \;=\; (S_\ell \odot \Theta_\ell)\, M_\ell\, x_\ell + b_\ell,
\end{equation}
where $M_\ell$ is the relaxed (doubly-stochastic) matrix during the soft phase and the hard permutation $\Pi_\ell$ after hardening. The corresponding backward pass follows directly from Eq.~\ref{eq:dst-forward}:

With $W_\ell = S_\ell \odot \Theta_\ell$ and output gradient $g_y = \partial \mathcal{L}/\partial y_\ell$, gradients are
\begin{align}
    \frac{\partial \mathcal{L}}{\partial \Theta_\ell}
    &\;=\; S_\ell \odot \big(g_y\, (M_\ell x_\ell)^\top\big),
    \label{eq:dst-back-w}\\
    \frac{\partial \mathcal{L}}{\partial M_\ell}
    &\;=\; W_\ell^\top g_y\, x_\ell^\top
    \;+\; \lambda_t \frac{\partial P(M_\ell)}{\partial M_\ell}.
    \label{eq:dst-back-m}
\end{align}
Inactive weights ($S_\ell = 0$) receive no gradient. After hardening, $\partial \mathcal{L}/\partial M_\ell$ is no longer computed, and the backward pass
reduces to that of the structured-sparse method.

The gradient updates above modify the values of permutations but not the mask $S_\ell$, which is updated separately by the DST method's prune-and-grow rule, raising a design choice about which coordinate system to update in.
Every $\Delta T$ steps, the DST method's prune-and-grow rule (RigL, SRigL, PixelatedBFly, DSB, DynaDiag) updates $S_\ell$ under the fixed sparsity budget. A design choice is whether prune-and-grow operates on $W_\ell$ in original coordinates or on $W_\ell M_\ell$ in permuted (i.e., post-shuffle) coordinates. We use \emph{permuted coordinates}: importance and growth scores are computed after applying the current permutation, so prune-and-grow sees the same channel grouping that the structured GEMM will see at inference.

Now, the remaining schedule decision is when to stop optimizing $M_\ell$ and decode a hard permutation. Permutation learning runs for the first $T_{\mathrm{perm}}$ steps. We monitor the per-layer penalty $P(M_\ell)$; once $P(M_\ell) < \delta$ for a layer, we decode that layer's permutation by solving for the hard permutation closest
to the current soft matrix. Equivalently, picking the $\Pi$ whose $1$s sit on the largest entries of $M_{\ell,k}$. We solve this linear assignment problem by the Hungarian algorithm~\citep{kuhn1955hungarian}:
\begin{equation}
\label{eq:hungarian}
    \Pi_{\ell,k} \;=\; \arg\max_{\Pi \in \mathcal{P}_{B_\pi}} \langle \Pi, M_{\ell,k} \rangle.
\end{equation}
replace $M_\ell$ by $\Pi_\ell$, and freeze the permutation for the remainder of training. Different layers can harden at different times. Mask updates continue on already-hardened layers, still in permuted coordinates. The permutation is now fixed but still applied. 

Tying these pieces together, within each training step, we (i) forward with current $\Theta$, $S$, $M$; (ii) compute $\mathcal{L}_{\mathrm{task}} + \lambda_t \sum P(M_\ell)$; (iii) backward and update $\Theta$ and unhardened $M$; (iv) project unhardened M back to the doubly-stochastic set and check the hardening trigger ($\delta$); (v) every $\Delta T$ steps, run prune-and-grow on $\Theta$ in permuted coordinates.

\subsection{2:4 Pruning with \MethodName}
\label{sec:method:prune}

For the pruning setting, we build directly on PermLLM~\citep{zou2025permllm}. PermLLM casts post-training $N{:}M$ pruning as learning a per-layer block-diagonal permutation that minimizes a cosine reconstruction loss against the dense model on calibration data. From PermLLM we inherit the pruning-in-shuffled-coordinates formulation, the two straight-through estimators~\cite{}: one for the Hungarian decoding~\cite{} of $M_\ell$, one for the hard top-$N$-of-$M$ mask selection, and the final-hardening
procedure in which $M_\ell$ is decoded once at the end and discarded. Our contribution in this setting is narrower: we replace PermLLM's Sinkhorn-based soft permutation with the AutoShuffleNet $\ell_1{-}\ell_2$ relaxation already used in DST (Sec.~\ref{sec:method:dst}). Concretely, the total objective becomes
\begin{equation}
\label{eq:prune-loss}
    \mathcal{L} \;=\; \mathcal{L}_{\mathrm{rec}}
    \;+\; \lambda_t \sum_\ell P(M_\ell),
\end{equation}
where $\mathcal{L}_{\mathrm{rec}}$ is the cosine reconstruction loss of \citep{zou2025permllm} between the sparse and dense outputs on a calibration set $\mathcal{C}$, and $P(M_\ell)$ is the row/column $\ell_1{-}\ell_2$ penalty from Eq.~\ref{eq:permloss}. Pretrained weights $W_\ell$ are frozen; only the per-layer permutations $M_\ell$ are optimized.

\subsection{Inference}
\label{sec:method:inf}
During inference, each learned permutation, $P_i$, is fixed and hence we make sure that permutations don't result in a separate matrix multiply operation. To do so, we precompute an index map $\ell:[d]\to[d]$ so the structured-sparse GEMM reads its input at permuted addresses, leaving the kernel, FLOP count, and sparse layout of $W$ unchanged from the baseline. Fully folding $\Pi$ into $W$ to eliminate all permutation cost is, however, structurally difficult so a runtime reorder is unavoidable, and a small residual overhead from imperfect memory-access
coalescing remains. We give the per-layer re-indexing, its backward pass, and
the absorption argument in full in \Apx{apx:inf}.

\section{Experimental Setup}
\label{sec:exp:setup}

We evaluate \MethodName in two regimes: dynamic sparse training (DST; \Sect{sec:setup_dst}) and one-shot semi-structured pruning of pretrained LLMs (\Sect{sec:setup_pruning}).

\subsection{Dynamic Sparse Training}
\label{sec:setup_dst}

\paragraph{Models and datasets.}
ViT-B/16 and ViT-L/16~\citep{dosovitskiy2020image} and MLP-Mixer-S/16~\citep{tolstikhin2021mlp} (Main text focuses on ViT-B/16; ViT-L/16 and Mixer-S/16 results appear in \Apx{app:downstream_full}) on ImageNet-1K~\citep{deng2009imagenet}; GPT-2 Small and GPT-2 Medium~\citep{radford2019language} on WikiText-103~\citep{merity2016pointer}. The sparsified layers in each model are listed in \Apx{sec:apx:layerDetails}.

\paragraph{Sparsity levels.}
$s \in \{60, 70, 80, 90, 95\}\%$ for vision and
$s \in \{40, 50, 60, 80, 90\}\%$ for language.

\paragraph{Baselines.}
(i) Dense ($s{=}0$); (ii) unstructured DST: RigL~\citep{evci2020rigging},
SET~\citep{mocanu2018scalable}, MEST~\citep{yuan2021mest}, CHT~\citep{zhang2024epitopological}, CHTs~\citep{zhang2025brain};
(iii) structured DST: SRigL~\citep{lasby2023dynamic}, DSB~\citep{jiang2022exposing}, DynaDiag~\citep{tyagi2025dynamic};
(iv) structured SST: Pixelated Butterfly~\citep{dao2021pixelated}.
We additionally augment each structured baseline with either \emph{learned} (\MethodName) or \emph{random} permutations fixed at initialization.

\paragraph{Metrics.}
Top-1 accuracy on ImageNet-1K (3 seeds) and perplexity on WikiText-103
(2 seeds), plus end-to-end training and inference wall-clock
(\Sect{sec:results:inftime}).

\subsection{One-Shot Semi-Structured Pruning}
\label{sec:setup_pruning}

\paragraph{Models and data.}
Qwen2.5-3B/7B~\citep{qwen2024qwen2} and
LLaMA2-7B~\citep{touvron2023llama}, calibrated on 128 random
1024-token C4 sequences~\citep{dodge2021documenting} following
\citet{frantar2023sparsegpt,sun2023simple}. We evaluate WikiText-2
perplexity~\citep{merity2016pointer} and zero-shot accuracy on
HellaSwag, ARC-Easy, ARC-Challenge, and OpenBookQA via
\texttt{lm-evaluation-harness}~\citep{eval-harness}, under the 2:4
sparsity pattern.

\paragraph{Baselines.}
SparseGPT~\citep{frantar2023sparsegpt}, Wanda~\citep{sun2023simple},
and PermLLM~\citep{zou2025permllm}, our most direct comparison, as it
also learns permutations for N:M sparsity post-training.

\section{Results}
\label{sec:results}




\begin{table*}[t]
\centering
\caption{ImageNet-1K Top-1 accuracy of ViT-B/16 under different sparsity patterns and sparsity levels. Avg Gap reports the average relative accuracy drop from the best unstructured method averaged across all sparsity levels; \textbf{lower is better}. \MethodName consistently reduces this gap below both the structured baseline and the random-permutation control across all four base methods.}
\label{tab:vitb16_results_integrated}
\definecolor{padstgray}{gray}{0.93}
\definecolor{deltagreen}{HTML}{1F8A3D}
\resizebox{0.75 \textwidth}{!}{%
\begin{tabular}{lcccccc}
\toprule
\textbf{Method} 
& \multicolumn{5}{c}{\textbf{Accuracy (\%)}} 
& \textbf{Avg Gap (\%)} \\
\cmidrule(lr){2-6} 
& \textbf{60\%} 
& \textbf{70\%} 
& \textbf{80\%} 
& \textbf{90\%} 
& \textbf{95\%} 
& \textbf{(lower is better)} \\
\midrule
\textit{Dense} & \multicolumn{5}{c}{78.50} & -- \\
\midrule
CHT  & 79.78 & 79.37 & 79.06 & 77.66 & 71.68 & -- \\
CHTs & 79.88 & 79.38 & 79.05 & 77.54 & 71.61 & -- \\
\midrule

SRigL                            & 77.81 & 77.82 & 77.39 & 75.94 & 68.73 & 2.61 \\
\quad + Random                   & 77.95 & 77.81 & 77.31 & 75.69 & 68.74 & 2.65 \\
\rowcolor{padstgray}
\quad + \MethodName              & 78.04 & 78.02 & 77.83 & 76.16 & 69.24 & 1.84 \,\textcolor{deltagreen}{$\downarrow$\,0.77} \\
\midrule

PixelatedBFly                    & 78.04 & 77.90 & 77.31 & 73.89 & 62.52 & 4.80 \\
\quad + Random                   & 77.91 & 77.94 & 77.34 & 73.93 & 62.45 & 4.83 \\
\rowcolor{padstgray}
\quad + \MethodName              & 78.10 & 78.04 & 77.49 & 74.09 & 62.82 & 4.04 \,\textcolor{deltagreen}{$\downarrow$\,0.76} \\
\midrule

DSB                              & 77.98 & 77.85 & 76.26 & 72.89 & 64.17 & 4.89 \\
\quad + Random                   & 78.06 & 77.84 & 76.27 & 72.84 & 64.23 & 4.87 \\
\rowcolor{padstgray}
\quad + \MethodName              & 78.11 & 77.95 & 76.34 & 73.09 & 64.49 & 4.06 \,\textcolor{deltagreen}{$\downarrow$\,0.83} \\
\midrule

DynaDiag                         & 78.29 & 77.94 & 77.62 & 76.91 & 69.54 & 1.92 \\
\quad + Random                   & 78.21 & 77.92 & 77.67 & 76.93 & 69.54 & 1.92 \\
\rowcolor{padstgray}
\quad + \MethodName              & 78.53 & 78.26 & 77.85 & 77.19 & 70.12 & 0.97 \,\textcolor{deltagreen}{$\downarrow$\,0.95} \\
\bottomrule
\end{tabular}%
}
\end{table*}

\setlength{\tabcolsep}{3.2pt}
\renewcommand{\arraystretch}{1.05}

\begin{table*}[t]
\centering
\caption{Perplexity of GPT-2 Medium across sparsity levels and downstream task performance at 90\% sparsity. PPL Gap reports the average relative increase in perplexity over the unstructured sparse baseline RigL across sparsity levels; lower is better, and 0 means matching RigL. The green arrow reports the reduction in PPL Gap relative to the corresponding structured baseline. Last column reports the absolute percentage-point difference between the downstream average of each method and RigL; higher is better, and 0 means matching RigL. \MethodName consistently reduces the gap to RigL in perplexity and also narrows the downstream gap for each structured sparsity family.}
\label{tab:gpt_m_results}
\definecolor{padstgray}{gray}{0.93}
\definecolor{deltagreen}{HTML}{1F8A3D}
\definecolor{deltared}{HTML}{B00020}
\setlength{\tabcolsep}{3.2pt}
\renewcommand{\arraystretch}{1.05}
\resizebox{\textwidth}{!}{%
\begin{tabular}{lccccccrrrrrrr}
\toprule
\textbf{Method}
& \multicolumn{5}{c}{\textbf{Perplexity ($\downarrow$)}} 
& \textbf{PPL Gap}
& \multicolumn{7}{c}{\textbf{Downstream at 90\% Sparsity ($\uparrow$)}} \\
\cmidrule(lr){2-6}
\cmidrule(lr){8-14}
& \textbf{40\%} & \textbf{50\%} & \textbf{60\%} & \textbf{80\%} & \textbf{90\%} 
& \textbf{(\%)}
& \textbf{LAMB.} & \textbf{ARC-E} & \textbf{ARC-C} & \textbf{HellaS.} & \textbf{WinoG.} & \textbf{Avg} & \textbf{\% Gap} \\
\midrule
\textit{Dense} & \multicolumn{5}{c}{20.18} & --
& 55.48 & 49.07 & 21.50 & 33.31 & 53.12 & 42.50 & - \\
\midrule

RigL & 20.45 & 21.60 & 23.49 & 28.87 & 51.76 & 0.00
& 40.83 & 41.18 & 18.47 & 27.49 & 51.21 & 35.84 & - \\
\midrule

SRigL                  & 21.14 & 22.59 & 26.09 & 32.16 & 55.66 & 7.59
& 39.26 & 39.39 & 17.13 & 23.22 & 48.93 & 33.59 & \textcolor{deltared}{-2.25} \\
\quad + Random         & 21.19 & 22.55 & 26.01 & 32.19 & 55.69 & 7.57
& 33.28 & 31.02 & 12.97 & 20.17 & 41.67 & 27.82 & \textcolor{deltared}{-8.02} \\
\rowcolor{padstgray}
\quad + \MethodName    & 20.57 & 22.20 & 25.04 & 29.89 & 55.13 & 4.00 \,\textcolor{deltagreen}{$\downarrow$\,3.59}
& 40.11 & 40.73 & 18.29 & 25.48 & 49.79 & 34.88 & \textcolor{deltared}{-0.96} \\
\midrule

PixelatedBFly          & 20.86 & 22.49 & 25.45 & 34.24 & 56.09 & 8.29
& 34.37 & 36.51 & 15.17 & 22.74 & 47.54 & 31.27 & \textcolor{deltared}{-4.57} \\
\quad + Random         & 20.90 & 22.51 & 25.44 & 34.22 & 56.01 & 8.29
& 33.92 & 31.88 & 14.29 & 21.84 & 45.89 & 29.56 & \textcolor{deltared}{-6.28} \\
\rowcolor{padstgray}
\quad + \MethodName    & 20.69 & 22.23 & 25.31 & 33.19 & 55.71 & 6.89 \,\textcolor{deltagreen}{$\downarrow$\,1.40}
& 36.26 & 37.86 & 16.36 & 22.97 & 48.24 & 32.34 & \textcolor{deltared}{-3.50} \\
\midrule

DynaDiag               & 20.69 & 22.14 & 24.98 & 29.65 & 54.87 & 3.75
& 38.45 & 39.14 & 17.29 & 24.75 & 49.26 & 33.78 & \textcolor{deltared}{-2.06} \\
\quad + Random         & 21.65 & 22.67 & 25.17 & 30.39 & 54.81 & 5.83
& 32.39 & 37.30 & 16.49 & 22.92 & 45.45 & 30.91 & \textcolor{deltared}{-4.93} \\
\rowcolor{padstgray}
\quad + \MethodName    & 20.55 & 21.91 & 24.71 & 29.21 & 54.26 & 2.63 \,\textcolor{deltagreen}{$\downarrow$\,1.12}
& 39.79 & 41.04 & 17.89 & 26.97 & 50.79 & 35.30 & \textcolor{deltared}{-0.54} \\
\bottomrule
\end{tabular}%
}
\end{table*}

\begin{table*}[t]
\centering
\caption{WikiText-2 perplexity and zero-shot downstream performance for 2:4 sparse models. PPL on WikiText-2 (lower is better); downstream accuracies on four tasks (higher is better). $acc$ used for HellaSwag and ARC-Easy, $acc\textnormal{-}norm$ for ARC-Challenge and OpenBookQA. The Average column reports the mean over the four downstream benchmarks; Gap to Dense reports the absolute drop in average accuracy from the dense reference (lower is better). \MethodName$_{\text{Wanda}}$ produces the smallest gap to Dense on LLaMA-2 7B and Qwen-2.5 3B and matches PermLLM$_{\text{Wanda}}$~\citep{zou2025permllm} on Qwen-2.5 7B, demonstrating that learnable permutations close the dense--sparse gap at LLM scale even with frozen weights. \MethodName values are mean $\pm$ std over 3 calibration seeds; PermLLM$_{\text{Wanda}}$ values are from original sources.}
\label{tab:ppl_downstream_2to4}
\definecolor{padstgray}{gray}{0.93}
\definecolor{deltagreen}{HTML}{1F8A3D}
\small
\setlength{\tabcolsep}{4.2pt}

\newcommand{\sd}[1]{{\scriptsize$\pm$#1}}

\resizebox{\textwidth}{!}{%
\begin{tabular}{llc|cccccl}
\toprule
\textbf{Model} & \textbf{Method} 
& \textbf{PPL ($\downarrow$)} 
& \textbf{HellaSwag} 
& \textbf{ARC-E} 
& \textbf{ARC-C} 
& \textbf{OBQA} 
& \textbf{Avg ($\uparrow$)} 
& \textbf{Gap to Dense ($\downarrow$)} \\
\midrule

\multirow{6}{*}{LLaMA-2 7B}
& \textit{Dense}                  & $5.47$\sd{0.011}  & $57.13$\sd{0.023} & $76.30$\sd{0.013} & $43.26$\sd{0.024} & $31.60$\sd{0.017} & $52.07$\sd{0.022} & --- \\
\cmidrule(lr){2-9}
& SparseGPT                       & $11.12$\sd{0.022} & $44.11$\sd{0.014} & $64.14$\sd{0.025} & $31.31$\sd{0.018} & $24.20$\sd{0.022} & $40.94$\sd{0.019} & $11.13$ \\
& Wanda                           & $12.16$\sd{0.011} & $41.59$\sd{0.021} & $61.74$\sd{0.015} & $30.20$\sd{0.026} & $24.00$\sd{0.012} & $39.38$\sd{0.027} & $12.69$ \\
& Wanda + RandPerm                & $14.19$\sd{0.007} & $40.11$\sd{0.039} & $57.29$\sd{0.003} & $26.22$\sd{0.006} & $23.71$\sd{0.003} & $36.83$\sd{0.015} & $15.24$ \\
\cmidrule(lr){2-9}
& PermLLM$_{\text{Wanda}}$        & $9.39$            & $46.60$           & $65.49$           & $31.14$           & $26.20$           & $42.36$           & $9.71$ \\
\rowcolor{padstgray}
& \MethodName$_{\text{Wanda}}$    & $9.09$\sd{0.020}  & $46.91$\sd{0.019} & $70.36$\sd{0.017} & $37.84$\sd{0.024} & $27.23$\sd{0.013} & $45.56$\sd{0.018} & \textcolor{deltagreen}{$\mathbf{6.51}$} \\
\midrule

\multirow{6}{*}{Qwen-2.5 3B}
& \textit{Dense}                  & $9.01$\sd{0.037}  & $54.78$\sd{0.004} & $71.45$\sd{0.026} & $42.68$\sd{0.012} & $30.05$\sd{0.016} & $49.74$\sd{0.037} & --- \\
\cmidrule(lr){2-9}
& SparseGPT                       & $19.64$\sd{0.023} & $44.06$\sd{0.003} & $64.14$\sd{0.007} & $33.48$\sd{0.001} & $24.64$\sd{0.006} & $41.58$\sd{0.006} & $8.16$ \\
& Wanda                           & $20.41$\sd{0.015} & $40.69$\sd{0.004} & $62.97$\sd{0.005} & $31.59$\sd{0.016} & $24.44$\sd{0.014} & $39.92$\sd{0.007} & $9.82$ \\
& Wanda + RandPerm                & $24.71$\sd{0.014} & $30.07$\sd{0.029} & $50.02$\sd{0.011} & $27.64$\sd{0.004} & $20.14$\sd{0.004} & $31.96$\sd{0.012} & $17.78$ \\
\cmidrule(lr){2-9}
& PermLLM$_{\text{Wanda}}$        & $17.42$           & $45.74$           & $67.89$           & $35.72$           & $26.06$           & $43.85$           & $5.89$ \\
\rowcolor{padstgray}
& \MethodName$_{\text{Wanda}}$    & $15.74$\sd{0.027} & $47.86$\sd{0.042} & $68.43$\sd{0.029} & $35.91$\sd{0.018} & $25.93$\sd{0.022} & $44.53$\sd{0.027} & \textcolor{deltagreen}{$\mathbf{5.21}$} \\
\midrule

\multirow{6}{*}{Qwen-2.5 7B}
& \textit{Dense}                  & $7.74$\sd{0.091}  & $58.79$\sd{0.025} & $79.56$\sd{0.016} & $46.08$\sd{0.023} & $33.00$\sd{0.021} & $54.36$\sd{0.027} & --- \\
\cmidrule(lr){2-9}
& SparseGPT                       & $14.34$\sd{0.026} & $46.20$\sd{0.018} & $71.13$\sd{0.021} & $37.46$\sd{0.015} & $26.00$\sd{0.029} & $45.20$\sd{0.011} & $9.16$ \\
& Wanda                           & $24.44$\sd{0.016} & $40.60$\sd{0.027} & $67.17$\sd{0.016} & $33.45$\sd{0.028} & $25.40$\sd{0.018} & $41.66$\sd{0.024} & $12.70$ \\
& Wanda + RandPerm                & $26.29$\sd{0.027} & $34.21$\sd{0.024} & $57.33$\sd{0.001} & $29.71$\sd{0.009} & $21.07$\sd{0.004} & $35.58$\sd{0.001} & $18.78$ \\
\cmidrule(lr){2-9}
& PermLLM$_{\text{Wanda}}$        & $13.58$           & $47.30$           & $70.58$           & $38.13$           & $27.60$           & $45.90$           & $\mathbf{8.46}$ \\
\rowcolor{padstgray}
& \MethodName$_{\text{Wanda}}$    & $13.61$\sd{0.026} & $47.12$\sd{0.026} & $70.34$\sd{0.024} & $38.34$\sd{0.015} & $27.47$\sd{0.029} & $45.82$\sd{0.024} & $8.54$ \\
\bottomrule
\end{tabular}%
}
\end{table*}

We evaluate \MethodName against two questions. First, does a single permutation per layer help close the gap between structured sparsity and unstructured/dense baselines, both during dynamic sparse training (\Sect{sec:result:dst}) and in post-training pruning of pretrained LLMs (\Sect{sec:res:prune})? Second, is the closing of gap driven by learning the permutation specifically, robust to its hyperparameters, and structurally consistent across the two regimes (\Sect{sec:results:abl}, \Sect{sec:results:shared_perm})?

\subsection{Closing the structured-unstructured gap during training}
\label{sec:result:dst}

We claim that with learned permutations the gap between the performance of structured and unstructured sparsities can be reduced. We show the advantages of learned permutations during DST by studying the gains in accuracy and perplexity (PPL) values for two modalities and across 4 structured sparse baseline methods.

\paragraph{Vision Experiments: }We evaluate two architectures for vision tasks on ImageNet-1K ~\citep{deng2009imagenet} dataset. We train two variants of Vision Transformers (ViTs)~\citep{dosovitskiy2020image}, ViT-B/16 and ViT-L, to study scalability aspects of sparsity. We also train Mixer-S/16 variant of MLP-Mixers~\citep{tolstikhin2021mlp} for our studies. We test all the sparse training methods (with and without permutations) at 60\%, 70\%, 80\%, 90\%, and 95\% sparsity in the targeted layers in the networks. Details of which layers are sparsified can be found in \Apx{sec:apx:layerDetails}.

\paragraph{Results: }\Tbl{tab:vitb16_results_integrated} shows the Top-1 accuracy on ImageNet-1K for ViT-B/16. Across sparsity levels, unstructured methods remain the strongest baselines, with CHT/CHTs achieving the best accuracy at each sparsity level. However, learned permutations consistently improve structured sparse training and reduce the gap to unstructured sparsity. Among structured methods, DynaDiag is the strongest baseline, with an average relative gap of 1.92\% from the best unstructured method; adding \MethodName reduces this gap to 0.97\%, the lowest among all structured variants. The gains are most visible at high sparsity: DynaDiag improves from 76.91\% to 77.19\% at 90\% sparsity and from 69.54\% to 70.12\% at 95\% sparsity with \MethodName. Similar trends are observed for other structured patterns: \MethodName improves SRigL, PixelatedBFly, and DSB over their non-permuted counterparts, reducing their average gaps from 2.61\%, 4.80\%, and 4.89\% to 1.84\%, 4.04\%, and 4.06\%, respectively.

\paragraph{Language Experiments: }For language tasks, we evaluate on the WikiText-103 dataset using two variants (Small and Medium) of GPT-2~\citep{radford2019language} at varying sparsities $s \in \{40\%, 50\%, 60\%, 80\%, 90\%\}$. We sparsify both the attention and the MLP layers of the model.

\paragraph{Results.}We show the results for the GPT2-Medium model in \Tbl{tab:gpt_m_results} (GPT2-Small results can be found in \Tbl{tab:gpt_s_results} in appendix). We report the average PPL obtained from \textit{two} runs. We also report downstream benchmark accuracy at 90\% sparsity, where the difference between unstructured, structured, and permutation-based methods is most informative. Across both model sizes, learned permutations consistently improve structured sparse training and reduce the gap to RigL, our unstructured baseline. This holds across all structured methods. On GPT2-Small, \MethodName reduces the average perplexity gap by 1.98 percentage points for SRigL, 0.80 for PixelatedBFly, and 0.77 for DynaDiag. On GPT2-Medium, the reductions are larger: 3.59, 1.40, and 1.12 percentage points, respectively. At 90\% sparsity, \MethodName also improves downstream accuracy for all structured methods on GPT2-Medium. 
\subsection{\MethodName based pruning of pretrained LLMs}
\label{sec:res:prune}
We next test whether the same permutation formulation also helps in pruning of pretrained LLMs, where weights are frozen and only the permutation is learned. As shown in \Tbl{tab:ppl_downstream_2to4}, \MethodName improves over the strongest non-permuted sparse baseline across all three models. It improves average downstream accuracy by 4.62 points on LLaMA-2 7B and 2.95 points on Qwen-2.5 3B, outperforming PermLLM$_{\text{Wanda}}$ by 3.20 and 0.68 points, respectively. On Qwen-2.5 7B, \MethodName is effectively tied with PermLLM$_{\text{Wanda}}$ within noise, with 45.82 average accuracy versus 45.90. 

\subsection{Learned Permutations in DST and Pruning}
\label{sec:results:shared_perm}

Having studied the effect of permutations on two settings, we want to understand if the eventually learned permutations have any similarities. To compare them, we measure how far each learned permutation is from the identity using $1 - \frac{|P-I|_F}{\sqrt{2N}}$, where $P$ is the learned hard permutation matrix and $N$ is its dimension. Lower values indicate permutations closer to the identity, while higher values indicate stronger reordering.

\begin{figure*}[t]
    \centering
    \begin{minipage}[t]{0.48\textwidth}
        \centering
        \includegraphics[width=\linewidth]{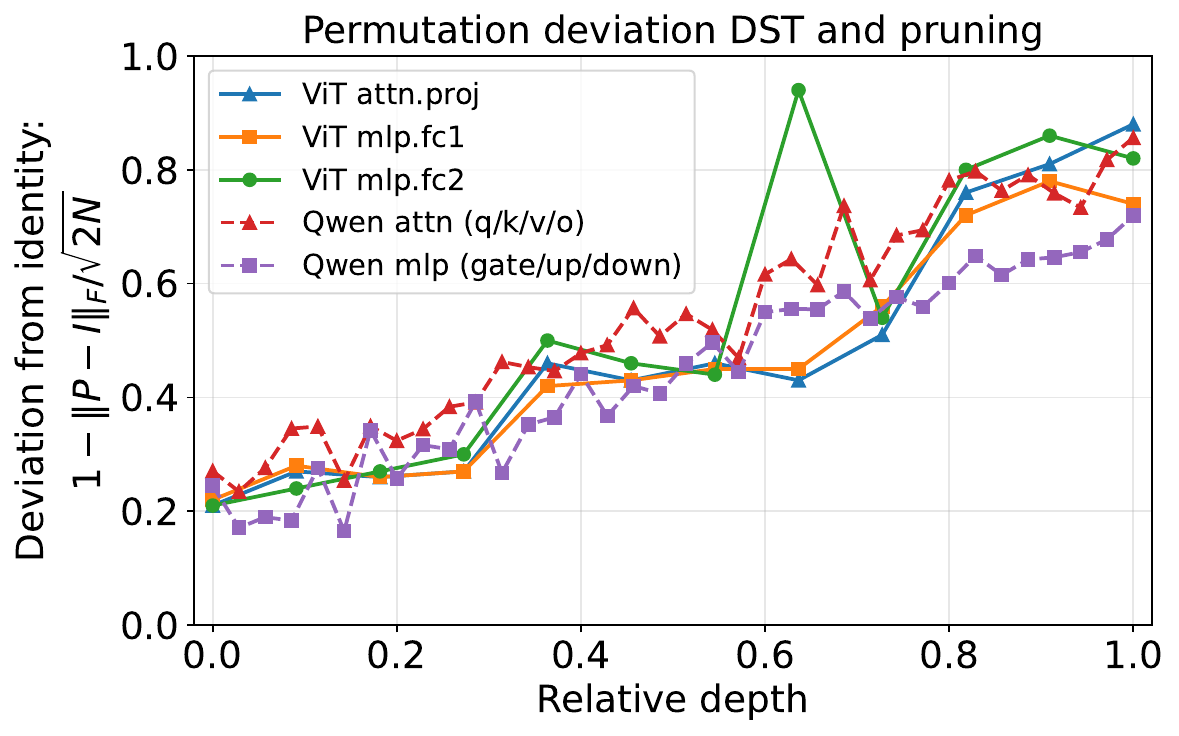}
        \caption{Permutation strength across depth in two regimes, where larger values on y-axis indicate stronger deviation from the identity. We show the results for a DST based ViT-B/16 at 90\% sparsity. We also show a Qwen-2.5 3B with 2:4 pruning. Both regimes show stronger permutations in later layers, and attention projections generally learn stronger permutations than MLP projections.}
        \label{fig:perm_cross_regime}
    \end{minipage}\hfill
    \begin{minipage}[t]{0.48\textwidth}
        \centering
        \includegraphics[width=\linewidth]{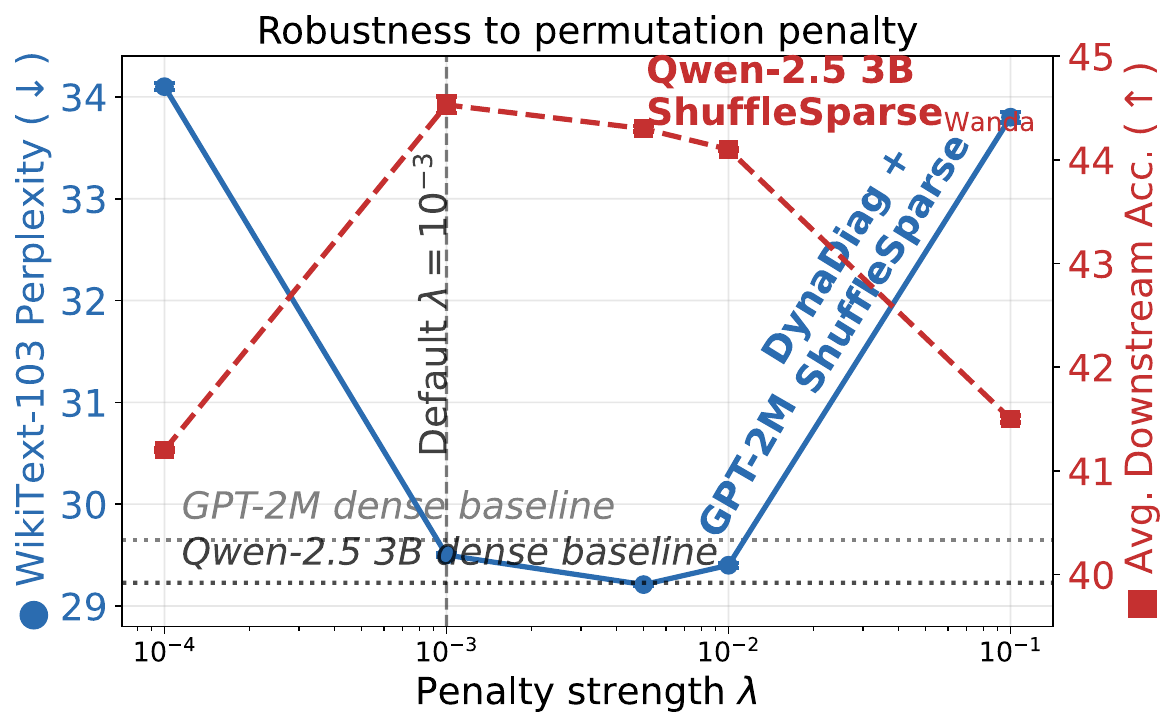}
        \caption{Sensitivity to the permutation penalty strength $\lambda$. We show results for a DST trained GPT-2 Medium with DynaDiag at 90\% sparsity. We also show a Qwen-2.5 3B with Wanda and 2:4 sparsity in the pruning regime. \MethodName is robust across an order of magnitude of $\lambda$, and the default value $\lambda=10^{-4}$ lies near the stable region in both regimes.}
        \label{fig:lambda_sweep}
    \end{minipage}
\end{figure*}

\Fig{fig:perm_cross_regime} shows this metric as a function of relative layer depth. For the DST setting, we use ViT-B/16 at 90\% sparsity and report the learned permutations for the attention output projection and the two MLP layers in each transformer block. For the pruning setting, we use Qwen-2.5 3B with 2:4 sparsity and report the average permutation strength across attention projections $(q/k/v/o)$ and MLP projections $(gate/up/down)$ in each block. In both cases, we first train the permutation module, project the soft permutation to a hard permutation, and then compute the distance-from-identity score for each layer.

The two settings show the same qualitative trend. Later layers learn stronger permutations than early layers, suggesting that deeper representations benefit more from feature reordering. In addition, attention projections show larger permutation deviation than MLP projections throughout most of the network. This pattern appears in both DST and post-hoc pruning, even though the optimization settings are different. The learned permutations are not identical across regimes, but their similar depth-wise behavior suggests that the module responds to common architectural structure rather than only regime-specific artifacts.

\subsection{Ablation study}
\label{sec:results:abl}

\paragraph{Random Permutations: }We test whether the gains come from learning useful permutations or simply from changing the channel order. We replace \MethodName with fixed random permutations and keep all other parts of the pipeline unchanged. Across vision, language-model training, and pruning results, random permutations can hurt the no-permutation baseline (or provide minimal gain), while learned permutations consistently improve it. The strongest example is Qwen-2.5 3B under 2:4 pruning: average downstream accuracy drops from 39.92 with Wanda to 31.96 with Wanda + RandPerm, but increases to 44.53 with \MethodName$_{\text{Wanda}}$. Similar trends appear in ViT-B/16 and GPT-2 experiments, where random permutations provide little or no gain, while learned permutations reduce the gap to the unstructured baseline. This shows that the improvements do not come from arbitrary reordering, but from learning task-specific permutations.

\paragraph{Sensitivity to penalty strength $\lambda$: }We study the sensitivity of \MethodName to the permutation penalty strength $\lambda$ in DST and pruning settings. \Fig{fig:lambda_sweep} shows that \MethodName is robust across an order of magnitude of $\lambda$. In the DST setting, GPT-2 Medium with DynaDiag at 90\% sparsity improves over the no-permutation baseline for a broad range of values, with the best perplexity obtained near the middle of the sweep. In the pruning setting, Qwen-2.5 3B with Wanda and 2:4 sparsity also remains above the no-permutation baseline across the stable range, with peak downstream accuracy around the same scale of $\lambda$. 

\paragraph{Training and inference overhead: }\MethodName adds a small overhead from learning and applying the permutation modules. In our inference measurements, this overhead ranges from 3.16\% to 8.69\% across models and sparsity settings. During training, the overhead also remains modest: at 95\% sparsity, \MethodName with DynaDiag takes only $1.23\times$ the dense training time. The full breakdown across methods and sparsity levels is provided in \Apx{apx:over}. Given the consistent accuracy and perplexity gains from learned permutations, we find this overhead acceptable in practice.



\section{Limitations}
\label{sec:limit}
We note three limitations of the current work. First, during DST the permutations are learnt at the cost of extra training time and memory overhead. This can be an issue at scale where the resources are constrained. We believe writing our kernel and coming up with a better permutation learning schedule can help us reduce the overhead significantly. Secondly, we study permutations for improving sparse neural networks. However, permuting inputs and outputs in weight matrices has already been shown to lead to improvements in quantization error (RPTQ~\citep{yuan2023rptq}, DuQuant~\citep{lin2024duquant}), and better KV cache compression~\citep{duanmu2024skvq}. We do not investigate whether a single learned permutation could fulfill several compression purposes but leave it as an open problem. Lastly, due to compute limits, we do not study permutation-learning dynamics for larger DST models. This remains an open problem, as learning overhead is expected to grow with model scale.
\section{Conclusion}
\label{sec:conclusion}
We show that \emph{learning one permutation matrix per layer} restores the expressivity lost in structured sparsity while preserving accelerator–friendly layouts. We back these claims with experimental results showing that training structured sparsity with the resulting \MethodName algorithm reduces the gap to unstructured-level accuracy while still realizing speed-ups during training and inference.  

\bibliography{example_paper}
\bibliographystyle{icml2026}


\appendix
\onecolumn
\appendix
 \begin{center}
     {\Large{\textbf{Appendix}}}
\end{center}



\section{Combinatorial Expressivity via Linear Regions}
\label{sec:analysis}

The question we would like to answer is: \textbf{Why is there an accuracy gap between the generalization performance of a structured vs an unstructured sparse method?}
Structured sparsity buys efficiency because accelerators can exploit regular patterns (blocks, $N{:}M$, diagonals).
Yet those same patterns limit which directions each layer can ``slice'' the input space along. 
Unstructured sparsity, and of course dense networks, do not impose such directional restrictions and typically attain higher accuracy at the same sparsity.
We make this precise by measuring expressivity via the \emph{number of linear regions} (NLR) of ReLU networks and by asking:
(i) how structure alone changes the classic depth-multiplicative region growth, and
(ii) how a single learned permutation per layer restores it when depth is sufficient.


\subsection{Theoretical Setup \& Intuition}
\label{sec:expr:setup}
We consider a depth-$L$ feed-forward ReLU network with layers $\ell=1,\dots,L$:
\[
z_\ell(x)=W_\ell\,a_{\ell-1}(x)+b_\ell,\qquad 
a_\ell(x)=\phi\!\big(z_\ell(x)\big),\qquad 
\phi(t)=\max\{t,0\},\qquad 
a_0(x)=x\in\mathbb{R}^{d_0}.
\]
Here $x\in\mathbb{R}^{d_0}$ is the input; $a_{\ell-1}(x)\in\mathbb{R}^{d_{\ell-1}}$ the activation at layer $\ell\!-\!1$; 
$W_\ell\in\mathbb{R}^{d_\ell\times d_{\ell-1}}$ and $b_\ell\in\mathbb{R}^{d_\ell}$ the weight matrix and bias; 
$z_\ell(x)\in\mathbb{R}^{d_\ell}$ the pre-activation; and $a_\ell(x)\in\mathbb{R}^{d_\ell}$ the post-ReLU activation (with $\phi$ applied elementwise). 
We write $n_\ell:=d_\ell$ for the width of layer $\ell$ and $(d_0,d_1,\dots,d_L)$ for the layer dimensions. 
The activation pattern is constant on convex polyhedra; each maximal such set is a \emph{linear region}.

\paragraph{Hyperplane arrangements \& expressivity via NLR.}
Each neuron contributes an affine hyperplane that “slices’’ the affine subspace propagated to its layer.  If, within that subspace, the hyperplanes are in \emph{subspace-general position} (SGP)—i.e., their restricted normals are in general position—standard arrangement counts apply. We denote by $\NLR(f)$ the total \emph{number of linear regions} of $f$. 
Classically, for dense ReLU networks with sufficiently wide layers, depth multiplies regions: each layer contributes a combinatorial factor that depends on the number of independent directions available, so $\NLR(f)$ grows multiplicatively with depth (Montúfar-style bounds \citep{montufar2014number}).

\paragraph{A generic lower-bound template.}
Let $k_\ell$ be the \emph{effective dimension} at layer $\ell$ (the number of independent row-normal directions realizable inside the current region). Under SGP,
\begin{equation}
\label{eq:nlr_master}
\NLR(f)\;\ge\;\prod_{\ell=1}^L \sum_{j=0}^{k_\ell} \binom{n_\ell}{j}.
\end{equation}
All architectural effects reduce to identifying $k_\ell$. 
To reason uniformly across settings, we track an \emph{accumulated span budget}
\begin{equation}
\label{eq:span_budget}
u_0:=0,\qquad u_\ell:=\min\{d_0,\,u_{\ell-1}+g_\ell\},
\end{equation}
where $g_\ell$ is the number of \emph{fresh} (linearly independent) directions that layer $\ell$ can inject beyond those already spanned. 
The effective dimension obeys
\begin{equation}
\label{eq:k_effective}
k_\ell=\min\{\,n_\ell,\;h_\ell\,\},\qquad \text{with } h_\ell\in\{u_{\ell-1},\,u_\ell\}
\end{equation}
depending on whether the newly available directions are usable inside the current region. 
Subsequent subsections instantiate \Eqn{eq:nlr_master}–\Eqn{eq:k_effective} for dense matrices and for unstructured sparsity before turning to structured sparsity and permutations later.

\subsection{Dense matrices: recovering classical multiplicative growth}
\label{sec:dense-multiplicative}
Dense layers impose no directional restriction: any normal in the ambient subspace can be realized. 
Thus, there are no “fresh’’ directions to accumulate ($g_\ell=0$) and the first layer already sees the full input subspace; we adopt the standard convention $u_0=d_0$. 
From \Eqn{eq:k_effective},
\begin{equation}
k_\ell=\min\{n_\ell,\,u_{\ell-1}\}=\min\{n_\ell,\,d_0\}\quad\text{for all }\ell,
\end{equation}
and plugging into \Eqn{eq:nlr_master} yields
\begin{equation}
\NLR(f)\;\ge\;\prod_{\ell=1}^L \sum_{j=0}^{\min\{n_\ell,d_0\}} \binom{n_\ell}{j}.
\end{equation}
If $n_\ell\ge d_0$ for all $\ell$, then $k_\ell=d_0$ layer-wise and each factor equals $\sum_{j=0}^{d_0}\binom{n_\ell}{j}$, giving the classical statement that \emph{depth multiplies regions}~\citep{montufar2014number}.

\subsection{Unstructured sparsity}
\label{sec:unstructured}
Unstructured sparsity does not impose intrinsic directional caps: with generic weights, the row normals of $W_\ell$ can span any $k\le \min\{n_\ell,d_{\ell-1}\}$ directions inside the current subspace. 
Therefore, as in the dense case, we take $g_\ell=0$ and $u_0=d_0$, which by \Eqn{eq:k_effective} gives
\begin{equation}
k_\ell=\min\{n_\ell,\,u_{\ell-1}\}=\min\{n_\ell,\,d_0\},
\end{equation}
and the lower bound \Eqn{eq:nlr_master} coincides with the dense bound:
\begin{equation}
\NLR(f)\;\ge\;\prod_{\ell=1}^L \sum_{j=0}^{\min\{n_\ell,d_0\}} \binom{n_\ell}{j}.
\end{equation}
\textit{Interpretation.} In the NLR lens, truly unstructured sparsity has the same depth-multiplicative expressivity as dense networks at a given width profile; any observed gap with structured sparsity arises from structural caps that reduce $k_\ell$ (analyzed in later sections) rather than from sparsity alone.

\subsection{Structured Sparsity Without Mixing}
\label{sec:structured-no-perm}
We now turn to \emph{structured} (axis-aligned) sparsity \emph{without} any re-orientation across layers (e.g., no permutations/mixers). 
Let $\mathcal{A}_\ell\subset\mathbb{R}^{d_{\ell-1}}$ be the set of admissible row-normal directions at layer $\ell$ induced by the fixed pattern (diagonals, bands, blocks, or tied $N{:}M$ groups). 
Define the \emph{directional rank cap} at layer $\ell$ by
\begin{equation}
r_\ell:=\dim\!\big(\mathrm{span}(\mathcal{A}_\ell)\big)\ \le\ d_{\ell-1}.
\end{equation}
For the axis-aligned families we study, the orientation and admissible directions are the same at every depth, so $r_\ell=r_{\mathrm{struct}}$ is \emph{constant} across layers. 
Set $s\;:=\;\min\{d_0,\,r_{\mathrm{struct}}\}$. Because the admissible directions lie in the same $r_{\mathrm{struct}}$-dimensional coordinate subspace at every layer, the first layer can realize at most $s$ independent directions and no fresh directions can be injected later. 
Using the global template \Eqn{eq:nlr_master}, this yields
\begin{equation}
\label{eq:structured-no-perm-lb}
\NLR(f)\ \ge\ \prod_{\ell=1}^L \sum_{j=0}^{\min\{n_\ell,s\}} \binom{n_\ell}{j}.
\end{equation}
\textit{What this means.} Unlike dense or unstructured layers, $k_\ell$ is uniformly capped by $s$ independently of depth, so the per-layer factor in \Eqn{eq:nlr_master} is bounded and depth-multiplicative growth stalls.

\paragraph{Instantiations of $r_{\mathrm{struct}}$ for each sparsity structure:}
For each sparsity structure, we instantiate $r_{\mathrm{struct}}$ as follows: for \textbf{Diagonal-$K$}, $r_{\mathrm{struct}}=K$, hence $s=\min\{d_0,K\}$; for \textbf{Block-$B$}, $r_{\mathrm{struct}}=B$, hence $s=\min\{d_0,B\}$; and for \textbf{N:M} with a tied group template, $r_{\mathrm{struct}}=\alpha d_0$ with $\alpha=N/M$, hence $s=\alpha d_0$. Substituting each resulting $s$ into \Eqn{eq:structured-no-perm-lb} gives the corresponding lower bound on $\NLR(f)$.

\subsection{Structured Sparsity With Mixing}
\label{sec:structured-perm}
We now allow a \emph{per-layer re-orientation}—a learned permutation or, more generally, any full-rank \emph{mixer}—applied before the axis-aligned mask. 
Such mixing prevents successive layers from aligning to the same small coordinate subspace, so each structured layer can contribute up to $r_{\mathrm{struct}}$ \emph{fresh} independent directions until the input dimension $d_0$ is saturated.

\paragraph{Generic consequence of mixing.}
Let the span budget evolve as $u_0:=0$ and $u_\ell=\min\{d_0,\, u_{\ell-1}+r_{\mathrm{struct}}\}$, and use the effective-dimension cap $k_\ell=\min\{n_\ell,u_\ell\}$ in the master template \Eqn{eq:nlr_master}. 
This yields the mixing-enabled lower bound
\begin{equation}
\label{eq:structured-with-perm-lb}
\NLR(f)\ \ge\ \prod_{\ell=1}^L \sum_{j=0}^{\min\{n_\ell,\,u_\ell\}} \binom{n_\ell}{j},
\qquad u_\ell=\min\{d_0,\,u_{\ell-1}+r_{\mathrm{struct}}\}.
\end{equation}
\textit{Meaning.} Each layer injects $r_{\mathrm{struct}}$ new independent directions; the span budget grows additively and the dense per-layer factor is recovered after a short warm-up of
\begin{equation}
L_{\mathrm{overhead}}=\Big\lceil \frac{d_0}{r_{\mathrm{struct}}}\Big\rceil\ \text{layers}.
\end{equation}

The recipe above applies unchanged to each axis-structured family by substituting its $r_{\mathrm{struct}}$: for \textbf{Diagonal-$K$}, $r_{\mathrm{struct}}=K\ \Rightarrow\ u_\ell=\min\{d_0,\,u_{\ell-1}+K\}$ and $L_{\mathrm{overhead}}=\lceil d_0/K\rceil$; for \textbf{Block-$B$}, $r_{\mathrm{struct}}=B\ \Rightarrow\ u_\ell=\min\{d_0,\,u_{\ell-1}+B\}$ and $L_{\mathrm{overhead}}=\lceil d_0/B\rceil$; and for \textbf{N:M}, $r_{\mathrm{struct}}=\alpha d_0$ with $\alpha=N/M\ \Rightarrow\ u_\ell=\min\{d_0,\,u_{\ell-1}+\alpha d_0\}$ and $L_{\mathrm{overhead}}=\lceil M/N\rceil$.
With a single mixer per layer, axis-structured sparsity thus regains dense-like, depth-multiplicative expressivity at the same widths after an explicit, structure-dependent overhead.

\paragraph{Why permutations (and what mixing suffices)?}
Any \emph{full-rank} per-layer mixer that varies across depth suffices for the theory above (e.g., permutations, co-prime stride shuffles, S-random interleavers, or very sparse full-rank transforms such as butterfly-style). 
We emphasize \emph{permutations} because they are (i) parameter-free and invertible, (ii) cheap and friendly to accelerator memory layouts, and (iii) preserve axis-structure in $W_\ell$, allowing the same efficient kernels at inference.
Empirically, learned permutations tend to outperform fixed random shuffles; the framework, however, only requires full rank and depth variation—not learnability per se. We discuss the practical consequences and predictions of our theory in detail in ~\Apx{sec:practical-consequences}.

\section{Mapping density to pattern parameters}
\label{app:sparsity-mapping}
Given a target per-layer density $\delta\in(0,1]$ and input size $n_{\mathrm{in}}$, we choose the smallest feasible integers so that per-row nnz $\approx \delta\,n_{\mathrm{in}}$:
\[
K=B=\mathrm{round}(\delta\,n_{\mathrm{in}}),\qquad
2b{+}1=\text{nearest odd to }\delta\,n_{\mathrm{in}},\qquad
\alpha=\tfrac{N}{M}=\delta \ \ (\text{tied }N{:}M).
\]
For non-cyclic bands/diagonals, use wrap-around (or adjust a few edge rows by $\pm 1$ nnz) so the total nnz matches the target.  
In our ViT-L/16 surrogate at $\delta=0.05$:  
$n_{\mathrm{in}}{=}1024 \Rightarrow K{=}B{=}51,\ 2b{+}1{=}51$;  
$n_{\mathrm{in}}{=}4096 \Rightarrow K'{=}B'{=}205,\ 2b'{+}1{=}205$;  
tied $N{:}M$ uses $\alpha=0.05$ throughout.

\section{Worked-example calculations (details)}
\label{app:calc-details}
Using the master bound \eqref{eq:nlr_master} and span update $u_\ell=\min\{d_0,u_{\ell-1}+r_{\mathrm{struct}}(n_{\mathrm{in},\ell})\}$ with $d_0=1024$, the alternating widths yield per-block gain
\[
r_{\mathrm{pair}}=r_{\mathrm{struct}}(1024)+\min\{r_{\mathrm{struct}}(4096),\,d_0\}=51+205=256.
\]
Thus $u_{2t}=\min\{1024,\,256\,t\}$ and dense-like factors are guaranteed once $u_{2t}=1024$, i.e., $t=\lceil 1024/256\rceil=4$ blocks ($\approx 8$ layers). 
Without mixing, $u_\ell\equiv 51$ and the per-layer factor remains strictly below dense for all depth.

\section{Practical Consequences \& Predictions of our theory}
\label{sec:practical-consequences}
We now analyze the impact of our theory via a concise worked example where \emph{all} structured families operate at (approximately) the same sparsity level. 
To stay within our ReLU/MLP framework, we use an MLP surrogate whose layer sizes are motivated by ViT-L/16, which has $24$ encoder blocks (307M params). Each block contains a two-layer FFN with $d_{\text{model}}{=}1024$ and $d_{\text{ff}}{=}4096$. 
We analyze the MLP that stacks these FFN layers in order,
\[
\underbrace{1024 \to 4096 \to 1024}_{\text{block 1 FFN}}
\ \to\ 
\underbrace{1024 \to 4096 \to 1024}_{\text{block 2 FFN}}
\ \to\ \cdots\ (\text{$24$ blocks}),
\]
yielding a depth-$L{=}48$ ReLU-MLP used solely to set widths for the master bound \eqref{eq:nlr_master}.
At \textbf{95\% sparsity} (density $\delta{=}0.05$), the structural caps (Sec.~\ref{sec:structured-no-perm}) are
\[
r_{\mathrm{struct}}(1024)=51,\qquad r_{\mathrm{struct}}(4096)=205,
\]
for Diagonal-$K$, Banded-$b$ (with $2b{+}1$ odd), Block-$B$, and tied $N{:}M$ with $\alpha=N/M=0.05$ (mapping in App.~\ref{app:sparsity-mapping}). 
\emph{Consequences.} 
Without mixing, all axis-structured families stall at $k_\ell\!\le\!51$ (no multiplicative growth). 
With one mixer per layer (permutations suffice), the span budget grows additively; because widths alternate $1024\!\leftrightarrow\!4096$, each block contributes $51{+}205=256$ fresh directions toward $d_0{=}1024$, so dense-like per-layer factors are guaranteed after
\[
\big\lceil 1024/256 \big\rceil = 4\ \text{blocks} \ (\approx 8\ \text{layers}).
\]
In this setting, the structured families share the \emph{same} catch-up point (4 blocks) when mixing is used.

\subsection{Combinatorial Expressivity Example}
\label{sec:apdx:comb}
\label{subsec:worked}
Take $d_0=4$, equal widths $n_\ell=8$, and $L=3$ layers.
\begin{enumerate}
\item \textbf{Dense / Unstructured (RigL/SET-like)}: $k_\ell=\min\{n_\ell,d_0\}=4$ at every layer.
Per-layer factor $\sum_{j=0}^4\binom{8}{j}=1+8+28+56+70=163$.
Hence $NLR\ge 163^3$.
\item \textbf{Block-$B$ without permutations}, $B=2$: 
$k_\ell=\min\{n_\ell,B\}=2$ at every layer (no new directions with depth).
Per-layer factor $1+8+28=37$.
Hence $NLR\ge 37^3$.
\item \textbf{Block-$B$ with a learned permutation each layer}, $B=2$ ($r_s=2$): 
$u_0=0\!\to\!u_1=2$ and $u_2=4$; thereafter $u_\ell=4$.
Per-layer factors: layer 1 gives $37$, layers 2 and 3 give $163$ each.
Thus $NLR\ge 37\cdot 163\cdot 163$.
After a one-layer overhead, the per-layer factor \emph{matches dense}.
\end{enumerate}
This concretely illustrates the phenomenon: structure alone stalls multiplicative growth; adding permutations restores it after a short, explicit warm-up in depth.

\begin{table}[ht!]
\centering
\small
\caption{\textbf{Lower bounds summary} (instantiate \eqref{eq:nlr_master}). Here $d_0$ is the input dimension, $n_\ell$ the width, and $r_s\in\{K,2b{+}1,B\}$ the per-layer structural cap for diagonal/banded/block. For tied $N{:}M$, $\alpha=N/M$.}
\label{tab:lb_summary}
\begin{tabular}{lccc}
\toprule
\textbf{Setting} & \textbf{Effective $k_\ell$} & \textbf{Span recursion $u_\ell$} & \textbf{Depth overhead} \\
\midrule
Dense & $\min\{n_\ell,d_0\}$ & $u_\ell=d_0$ & 0 \\
Unstructured DST (free masks) & $\min\{n_\ell,d_0\}$ & $u_\ell=d_0$ & 0 \\
$N{:}M$ (free supports) & $\min\{n_\ell,d_0\}$ & $u_\ell=d_0$ & 0 \\
$N{:}M$ (tied template) & $\min\{n_\ell,\alpha u_{\ell-1}\}$ & $u_\ell=u_{\ell-1}$ & -- (stalls) \\
Diagonal-$K$ (no perm) & $\min\{n_\ell,K\}$ & $u_\ell=\min\{d_0,K\}$ & -- (stalls) \\
Banded-$b$ (no perm) & $\min\{n_\ell,2b{+}1\}$ & $u_\ell=\min\{d_0,2b{+}1\}$ & -- (stalls) \\
Block-$B$ (no perm) & $\min\{n_\ell,B\}$ & $u_\ell=\min\{d_0,B\}$ & -- (stalls) \\
Diagonal-$K$ + permutation & $\min\{n_\ell,u_\ell\}$ & $u_\ell=\min\{d_0,u_{\ell-1}+K\}$ & $\lceil d_0/K\rceil$ \\
Banded-$b$ + permutation & $\min\{n_\ell,u_\ell\}$ & $u_\ell=\min\{d_0,u_{\ell-1}+2b{+}1\}$ & $\lceil d_0/(2b{+}1)\rceil$ \\
Block-$B$ + permutation & $\min\{n_\ell,u_\ell\}$ & $u_\ell=\min\{d_0,u_{\ell-1}+B\}$ & $\lceil d_0/B\rceil$ \\
\bottomrule
\end{tabular}
\end{table}
\begin{algorithm}[t]
\caption{\MethodName permutation learning over a pretrained LLM}
\label{alg:prune}
\begin{algorithmic}[1]
\Require frozen weights $\{W_\ell\}$, calibration set $\mathcal{C}$,
activation scales $\{s_\ell\}$, permutation params $\{M_\ell\}$, total steps $T$
\For{$t = 1, \ldots, T$}
    \State $\lambda_t \gets \lambda_{\max}\,\min(1, t / T_{\mathrm{perm}})$
    \For{each pruned layer $\ell$}
        \State $\Pi_\ell \gets \textsc{Hungarian}(M_\ell)$;\quad
               $\bar{M}_\ell \gets \mathrm{sg}(\Pi_\ell - M_\ell) + M_\ell$
        \State $\tilde{A}_\ell \gets |W_\ell|\,\mathrm{diag}(s_\ell)\,\bar{M}_\ell$
        \State Compute $\tilde{Z}_\ell$ (hard top-$N$-of-$M$);
               $\hat{Z}_\ell$ (soft surrogate)
    \EndFor
    \State Forward $f_{\mathrm{sparse}}$ on a batch from $\mathcal{C}$ 
    \State $\mathcal{L} \gets \mathcal{L}_{\mathrm{rec}}
                        + \lambda_t \sum_\ell P(M_\ell)$
    \State Backprop and update $\{M_\ell\}$;\quad project to Birkhoff polytope
\EndFor
\State \textbf{Final hardening:} for each $\ell$, set
$\Pi_\ell \gets \textsc{Hungarian}(M_\ell)$, recompute $N{:}M$ masks under $\Pi_\ell$,
apply masks to $W_\ell$, store index map $\ell_\ell$ (Sec.~\ref{sec:method:inf}),
discard $M_\ell$.
\end{algorithmic}
\end{algorithm}

\section{Permutation Learning Schedule \& Overhead}
\label{sec:apx:permover}

\begin{figure}[ht!]
	\centering
	\includegraphics[width=\textwidth]{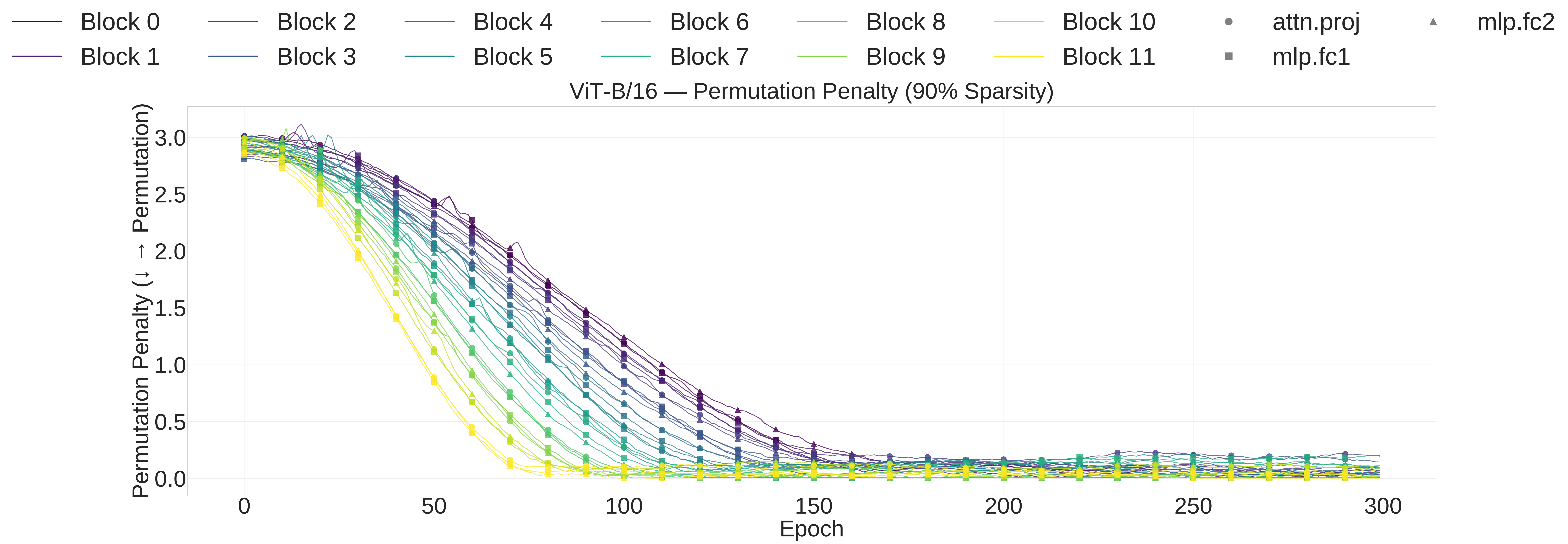}
	\caption{Tracking the loss of permutation matrices based on \Eqn{eq:permloss}. We plot the loss for every 10 training epochs.}
\label{fig:permloss}
\end{figure}

\subsection{Permutation Learning Schedule}
While training, we can either learn the permutations till the end of the training or find a way to stop early without losing out on optimal permutations. From our experiments, we find that it is useful to monitor the loss of the soft-permutation matrix. For our training, we set a threshold loss value $\delta$ at which point we go from a soft to a hard permutation matrix for that particular layer. This means that instead of carrying out a multiplication operation for that layer, we can instead do re-indexing, which is a much cheaper operation. 

\begin{figure}[t!]
	\centering
	\includegraphics[width=\textwidth]{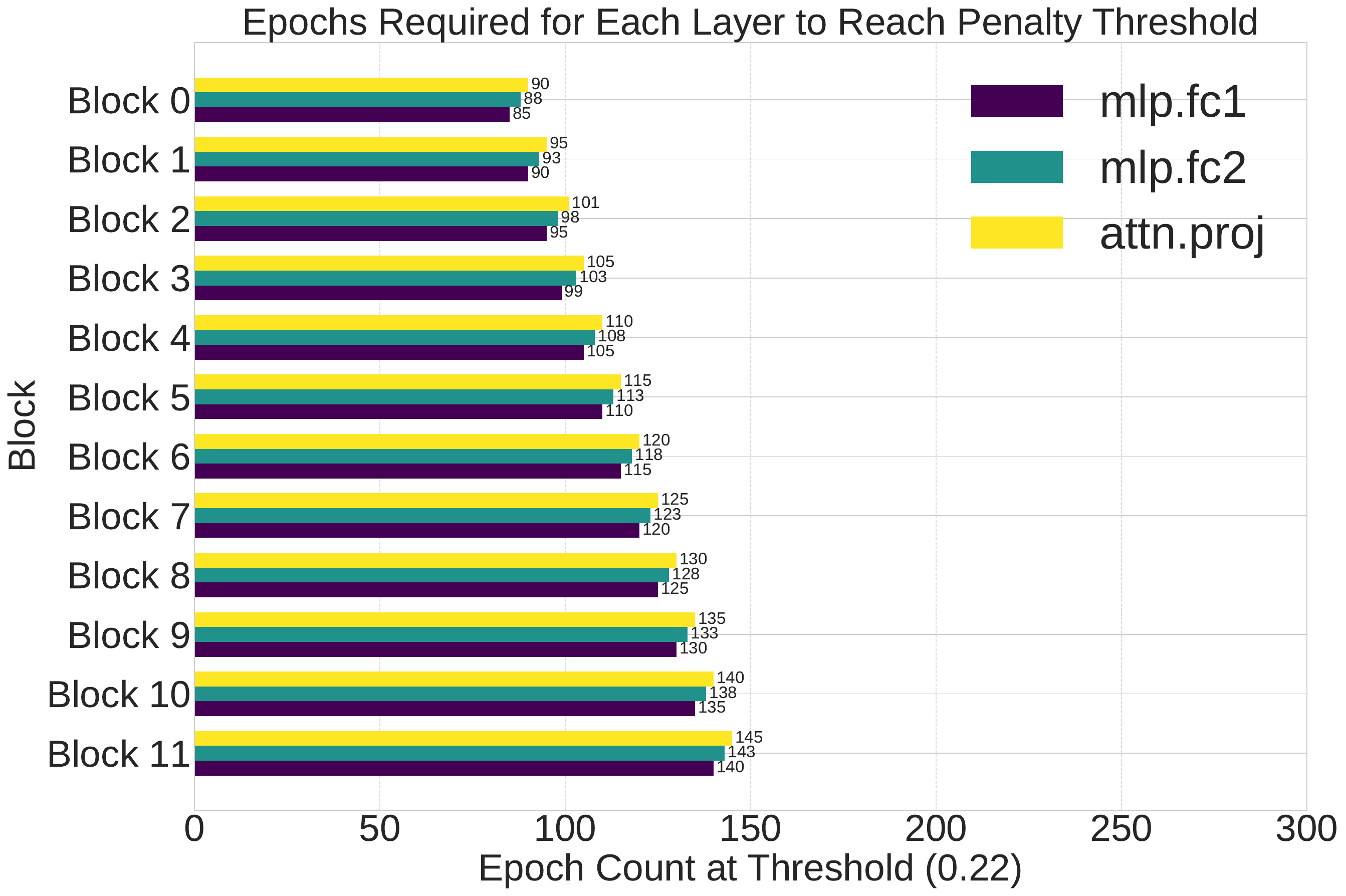}
	\caption{We show the epoch count of each layer in a ViT-B/16 network when the corresponding permutation loss hits the threshold. We can see that the cutoff epoch varies drastically acorss the network which we use to our advantage to reduce the training time.}
\label{fig:epcohthres}
\end{figure}

We show in \Fig{fig:permloss} the loss associated with each permutation matrix when training a ViT-B/16 network at 90\% sparsity with \method. We can see that the loss decreases drastically and saturates after the knee in the plot. Moreover, we see a clear trend that earlier permutation matrices take longer to reach this knee point. For the given experiment, we set the $\delta = 0.22$ and we show in \Fig{fig:epcohthres} when each layer reaches that threshold, and we stop training the permutation matrix corresponding to that layer.

\subsection{Permutation Learning Overheads and Additional Results}
\label{apx:over}
The tables quantify the training overhead associated with various permutation methods. Specifically, \Tbl{tab:mem_overhead_gpt2_diag} and \Tbl{tab:mem_overhead_gpt2_nm} detail the memory overhead in gigabytes and as a percentage for the GPT-2 Small model, using Diagonal and N:M sparsity, respectively. \Tbl{tab:mem_overhead_vitb_diag} presents a similar memory overhead analysis for the ViT-B/16 model, but at higher 90\% and 95\% sparsity levels. Finally, \Tbl{tab:overhead_gpt2_medium_diag} expands this analysis for the GPT-2 Medium model, showing the overhead for both training time (in hours) and memory. Across all tables, we compare permutation strategies such as AutoShufflePerm and the KaleidoscopePerm against their corresponding non-permuted baselines to clearly illustrate the computational trade-offs.

\begin{figure*}[ht!]
	\centering
	\includegraphics[width=0.85\textwidth]{figs/vitb16_times_side_by_side_with_legend.pdf}
	\caption{Impact of permutations on the training and inference time for a ViT-B/16 model. We observe that there is a 3.16\% - 8.69\% overhead related to permutations during inference for the obtained accuracy gains. Even for training, learning permutations increases the overall training time, but at higher sparsity levels, we still obtain speedup compared to a dense model for all structured sparsities. Training time for GPT models can be found in \Tbl{tab:overhead_gpt2_medium_diag}.}
\label{fig:infTime}
\end{figure*}

\section{Training and Inference Cost}
\label{sec:results:inftime}

\subsection{Inference: Re-indexing and the Absorbing Permutations During Inference}
\label{apx:inf}

After hardening, every learned $\Pi_\ell$ is fixed and the deployed model performs the same structured-sparse GEMMs as the corresponding baseline---no dense permutation matrix is ever multiplied. We make this concrete and then explain why some residual overhead is fundamental. 

\paragraph{Re-indexing.}
For each layer with hard permutation $\Pi$, we precompute an index map $\ell : [d] \to [d]$ such that $(\Pi x)_i = x_{\ell(i)}$. The structured-sparse GEMM
then reads $x$ at permuted addresses,
\begin{equation}
\label{eq:reindex}
    y_k \;=\; \sum_{i=1}^{d} W[k, i]\, x_{\ell(i)},
\end{equation}
with no $\Pi$ materialized at runtime. The same construction applies to both FFN projections (with maps $\ell_\uparrow, \ell_\downarrow$) and to the MHA output projection (with map $\ell_O$): the kernel-launch count, the FLOP count of the GEMM, and the structured layout of $W$ are all unchanged. 

\paragraph{Why some overhead is fundamental.}
A natural follow-up is why we cannot fold $\Pi$ into $W$ and reach zero overhead. The answer is structural: a column permutation of $W$ cannot be absorbed into $W$'s storage without destroying the layout that makes $W$ acceleratable. For $N{:}M$, the structure constrains nonzeros to occupy exactly $N$ positions in every group of $M$ contiguous columns; reordering columns scatters groupings and breaks the kernel's group invariant. For block sparsity, nonzeros live in contiguous tiles; permuting columns disperses tiles into non-contiguous layouts. For diagonal and banded structures, the structure \emph{is} the column-vs-row index relationship and cannot survive reordering. \emph{The permutation must therefore execute at runtime in some form.} Re-indexing eliminates the two largest costs: i)the extra kernel launch; ii) the explicit memory copy. But reindexing cannot eliminate the loss of memory-access coalescing on a memory-bound sparse GEMM. This residual cost is what we measure as $3.16$-$8.69\%$ overhead in \Apx{sec:results:inftime}.

Learning permutations means extra computation, and hence it is essential to understand the overheads associated with the same. We measure the inference and training wall clock times of various structured sparse training approaches with and without permutations. 

\paragraph{Setup. }We use the available libraries to accelerate the execution of structured sparsities mentioned in \Sect{sec:exp:setup}, and we use cuSparse~\citep{naumov2010cusparse} to execute all the unstructured sparsities. We obtain the execution times for SRigl~\citep{lasby2023dynamic} using their official codebase including the CUDA kernels. Whereas, for \textbf{DSB and PixelatedBFly}, we use the Triton based library package from PixelatedBFly~\citep{dao2021pixelated} to accelerate inference for both and training for PixelatedBFly (DSB lacks support to integrate the kernels with the training pipeline). And lastly, we use the provided CUDA kernels implemented to accelerate \textbf{DynaDiag's} execution on the GPUs.

\paragraph{Results.}In \Fig{fig:infTime}, we compare the inference and training times of models listed in \Sect{sec:exp:setup}. We see that using \MethodName, we can re-index the output of a layer instead of explicitly applying a permutation matrix, leading to an overhead of just up to \textbf{8.69\%} as compared to without permutation baselines at all sparsity levels. Even with the overhead, there is an inference speedup as high as 2.9$\times$ for DynaDiag at 90\% sparsity. Whereas, during training, we see an increase in training time for all structured sparse methods due to extra computation required for learning the permutations. We elaborate in \Apx{sec:apx:permover} overheads \& our approach to early stopping the permutation learning by tracking the loss and heuristically determining a stopping point. However, we can see that at higher sparsities, even with permutations, it is possible to get better training time than dense and unstructured sparse models. For example, at 95\% sparsity, the training times for SRigL, PBFly, DSB, and \method are 1.09$\times$, 1.18$\times$, 1.13$\times$, and 1.23$\times$ that of RigL, respectively.

\section{More Results}
\label{app:downstream_full}

\begin{table}[ht!]
\centering
\caption{Memory overhead of permutation methods for GPT-2 Small with Diagonal sparsity on WikiText-103. The overhead percentage is calculated relative to the \method method.}
\label{tab:mem_overhead_gpt2_diag}
\resizebox{\textwidth}{!}{%
\begin{tabular}{llcccc}
\toprule
\textbf{Model} & \textbf{Method} & \textbf{Memory (GB) @ 60\%} & \textbf{\% Overhead} & \textbf{Memory (GB) @ 80\%} & \textbf{\% Overhead} \\
\midrule
\multirow{5}{*}{\textbf{GPT-2 Small (\method)}}
& Unstructured & 133.64 & - & 132.12 & - \\
& \method & 137.32 & - (Baseline) & 132.82 & - (Baseline) \\
\cmidrule{2-6}
& \quad + FixedRandPerm & 139.21 & \textbf{+1.38\%} & 134.09 & \textbf{+0.96\%} \\
& \quad + \MethodName & 156.17 & \textbf{+13.73\%} & 145.41 & \textbf{+9.48\%} \\
& \quad + KaleidoscopePerm & 141.27 & \textbf{+2.88\%} & 137.43 & \textbf{+3.47\%} \\
\bottomrule
\end{tabular}%
}
\end{table}

\begin{table}[ht!]
\centering
\caption{Memory overhead of permutation methods for GPT-2 Small with SRigL on WikiText-103. The overhead percentage is calculated relative to the SRigL method.}
\label{tab:mem_overhead_gpt2_nm}
\resizebox{\textwidth}{!}{%
\begin{tabular}{llcccc}
\toprule
\textbf{Model} & \textbf{Method} & \textbf{Memory (GB) @ 60\%} & \textbf{\% Overhead} & \textbf{Memory (GB) @ 80\%} & \textbf{\% Overhead} \\
\midrule
\multirow{5}{*}{\textbf{GPT-2 Small}}
& Unstructured & 133.64 & - & 132.12 & - \\
& SRigL & 135.12 & - (Baseline) & 131.42 & - (Baseline) \\
\cmidrule{2-6}
& \quad + FixedRandPerm & 136.91 & \textbf{+1.32\%} & 132.65 & \textbf{+0.94\%} \\
& \quad + \MethodName & 148.23 & \textbf{+9.70\%} & 143.19 & \textbf{+8.96\%} \\
\bottomrule
\end{tabular}%
}
\end{table}

\begin{table}[ht!]
\centering
\caption{Memory overhead of permutation methods for ViT-B/16 with Diagonal sparsity on ImageNet-1K. The overhead percentage is calculated relative to the \method method.}
\label{tab:mem_overhead_vitb_diag}
\resizebox{\textwidth}{!}{%
\begin{tabular}{llcccc}
\toprule
\textbf{Model} & \textbf{Method} & \textbf{Memory (GB) @ 90\%} & \textbf{\% Overhead} & \textbf{Memory (GB) @ 95\%} & \textbf{\% Overhead} \\
\midrule
\multirow{6}{*}{\textbf{ViT-B/16 (Diag)}}
& Unstructured & 112.64 & - & 112.64 & - \\
& \method & 114.50 & - (Baseline) & 111.43 & - (Baseline) \\
\cmidrule{2-6}
& + \MethodName & 138.38 & \textbf{+20.86\%} & 135.01 & \textbf{+21.16\%} \\
& + KaleidoscopePerm & 120.67 & \textbf{+5.39\%} & 117.43 & \textbf{+5.38\%} \\
\bottomrule
\end{tabular}%
}
\end{table}

\begin{table}[ht!]
\centering
\caption{Time and memory overhead of permutation methods for GPT-2 Medium with Diagonal sparsity on WikiText-103. The overhead percentage is calculated relative to the \method method.}
\label{tab:overhead_gpt2_medium_diag}
\resizebox{\textwidth}{!}{%
\begin{tabular}{l l cccc cccc}
\toprule
& & \multicolumn{4}{c}{\textbf{Sparsity @ 60\%}} & \multicolumn{4}{c}{\textbf{Sparsity @ 80\%}} \\
\cmidrule(lr){3-6} \cmidrule(lr){7-10}
\textbf{Model} & \textbf{Method} & \textbf{Time (h)} & \textbf{\% Overhead} & \textbf{Memory (GB)} & \textbf{\% Overhead} & \textbf{Time (h)} & \textbf{\% Overhead} & \textbf{Memory (GB)} & \textbf{\% Overhead} \\
\midrule
\multirow{5}{*}{\textbf{GPT-2 Medium}}
& Unstructured & 36.43 & - & 152.34 & - & 36.32 & - & 152.32 & - \\
& \method & 33.21 & - (Base) & 149.64 & - (Base) & 29.82 & - (Base) & 142.82 & - (Base) \\
\cmidrule{2-10}
& \quad + FixedRandPerm & 33.39 & +0.54\% & 149.41 & -0.15\% & 29.23 & -1.98\% & 144.09 & +0.89\% \\
& \quad + \MethodName & 37.49 & \textbf{+12.88\%} & 156.17 & +4.36\% & 34.28 & \textbf{+14.95\%} & 145.41 & +1.81\% \\
& \quad + KaleidoscopePerm & 41.98 & \textbf{+26.41\%} & 155.27 & +3.76\% & 37.98 & \textbf{+27.36\%} & 137.43 & -3.77\% \\
\bottomrule
\end{tabular}%
}
\end{table}

\subsection{Experiment Details}
\label{sec:apdx:exp}
All experiments are conducted on the NVIDIA A100 GPUs with the following configuration:
\begin{itemize}
    \item Model: NVIDIA A100 40GB \\
     \item Memory: 40GB HBM2e \\
    \item Memory Bandwidth: $\sim$2.0 TB/s (higher than the 40GB version) \\
    \item TDP : 400W (PCIe: 300W) \\
    \item Peak FP32 Performance: $\sim$19.5 TFLOPS (same as 40GB) \\
     \item Peak FP16 Performance: $\sim$312 TFLOPS (same as 40GB) \\
\end{itemize}
\subsubsection{Datasets}
\begin{enumerate}

    \item \textbf{ImageNet-1K}~\cite{deng2009imagenet} covers 1,000 object classes, with 1.28M training, 50,000 validation, and 100,000 test images. Images are typically resized and cropped to $224 \times 224$ for processing.

    \item \textbf{WikiText-103}~\cite{merity2016pointer} comprises over 100 million tokens extracted from verified Wikipedia articles. It is significantly larger than other language datasets, such as Penn Treebank (PTB)~\cite{marcus1993building}.
\end{enumerate}

\begin{table}[ht!]
\centering
\caption{Configuration of the CIFAR10 and CIFAR100 experiments with MLPMixer.}
\label{tab:experiment_params_multicol}
\begin{tabular}{lc@{\hspace{2cm}}lc}
\toprule
\textbf{Parameter} & \textbf{Value} & \textbf{Parameter} & \textbf{Value} \\
\midrule
Adam $\beta_1$ & 0.9 & Hidden & 128 \\
Adam $\beta_2$ & 0.99 & (Initial LR, Final LR) & $(10^{-3}, 10^{-6})$ \\
AutoAugment & True & Label Smoothing & 0.1 \\
Batch Size & 128 & Layers & 8 \\
CutMix Probability & 0.5 & LR Scheduler & Cosine \\
CutMix $\beta$ & 1.0 & Optimizer & Adam \\
Dropout & 0.0 & Random Seed & 3407 \\
Epochs & 300 & Weight Decay & $5 \times 10^{-5}$ \\
Hidden\_C & 512 & Warmup & 5 epochs \\
Hidden\_S & 64 & & \\
\bottomrule
\end{tabular}
\end{table}
\hfill


\begin{table}[t]
\centering
\caption{Configuration of the ImageNet experiments with ViT-Base and MLPMixer. Here X represents any of the sparse training methods that train a ViT-Base network.}
\label{tab:imagenet_config}
\begin{tabular}{lcccccc}
\hline
\textbf{Model} & \textbf{Optimizer} & \textbf{Weight Decay} & \textbf{Learning Rate} & \textbf{Drop Path} & \textbf{Warmup/Epoch} \\
\hline
ViT-Base & AdamW & 0.05 & 0.001 & 0.1 & 5/300 \\
X-ViT-Base & AdamW & 0.05 & 0.001 & 0 & 5/300 \\
\hline
Mixer-Small & AdamW & 0.1 & 0.001 & 0.1 & 5/300 \\
X-Mixer-Small & AdamW & 0.1 & 0.001 & 0 & 5/300 \\
\hline
\end{tabular}
\end{table}

\begin{table}[!ht]
    \centering
    \caption{Configuration of the ImageNet experiments with ViT-Large and Huge.}
    \label{tab:ours_imnet1k_multicol}
    \begin{tabular}{lc@{\hspace{2cm}}lc}
    \toprule
    \textbf{Parameter} & \textbf{Value} & \textbf{Parameter} & \textbf{Value} \\
    \midrule
    Batch size              & 256                & Horizontal flip       & \cmark \\
    Optimizer               & AdamW              & Random Resized Crop (RRC)   & \cmark \\
    Learning Rate (LR)      & $3 \times 10^{-3}$ & Rand Augment          & \xmark \\
    LR decay                & cosine             & 3 Augment (ours)      & \cmark \\
    Weight decay            & 0.02               & LayerScale            & \cmark \\
    Warmup epochs           & 5                  & Mixup $\alpha$        & 0.8 \\
    Label smoothing $\varepsilon$ & 0.1                & Cutmix $\alpha$       & 1.0 \\
    Dropout                 & \xmark             & Erasing prob.         & \xmark \\
    Stochastic Depth        & \cmark             & ColorJitter           & 0.3 \\
    Repeated Aug            & \cmark             & Test crop ratio       & 1.0 \\
    Gradient Clipping       & 1.0                & Loss                  & BCE \\
    \bottomrule
    \end{tabular}
\end{table}

\begin{table}[t]
\centering
\caption{Configuration of the Wikitext-103 experiments GPT-2Small experiments.}
\label{tab:gpt2_config}
\begin{tabular}{lccccc}
\hline
\textbf{Model} & \textbf{Optimizer} & \textbf{Weight Decay} & \textbf{Learning Rate} & \textbf{Dropout} & \textbf{Warmup/Epoch} \\
\hline
GPT-2-Small & AdamW & 0.1 & 0.0001 & 0.1 & 5/100 \\
DynaDiag & AdamW & 0.1 & 0.0001 & 0.1 & 5/100 \\
\hline
\end{tabular}
\end{table}
\clearpage


\subsection{Raw Values}
\label{app:raw}

\begin{table*}[ht!]
\centering
\caption{Top-1 accuracy of structured sparse training methods at varying sparsities for a ViT-B/16 on ImageNet-1K. The results shown are from three runs for each data point. We see that there is no significant difference between the generalization performance of networks with learnt row and col permutation.}
\label{tab:abl:row}
\resizebox{\textwidth}{!}{
\begin{tabular}{l c c c c c c}
\toprule
\textbf{Method} & \textbf{Perm.} &\textbf{60\%} & \textbf{70\%} & \textbf{80\%} & \textbf{90\%} & \textbf{95\%} \\
\midrule
& & \multicolumn{5}{c}{\textit{dense accuracy = 78.5}} \\
\midrule
SRigL           & Col & 78.04 $\pm$ 0.011      & 78.02 $\pm$ 0.008      & 77.83 $\pm$ 0.009 & 76.16 $\pm$ 0.007 & 69.24 $\pm$ 0.008 \\
PixelatedBFly   & Col & 78.10 $\pm$ 0.005      & 78.04 $\pm$ 0.007      & 77.49 $\pm$ 0.007 & 74.09 $\pm$ 0.008 & 62.82 $\pm$ 0.006\\
DSB             & Col & 78.11 $\pm$ 0.009      & 77.95 $\pm$ 0.008      & 76.34 $\pm$ 0.005 & 73.09 $\pm$ 0.004 & 64.49 $\pm$ 0.004 \\
\method         & Col & 78.53 $\pm$ 0.007      & 78.26 $\pm$ 0.003      & 77.85 $\pm$ 0.005 & 77.19 $\pm$ 0.004 & 70.12 $\pm$ 0.007\\
\midrule
SRigL           & Row & 78.03 $\pm$ 0.007 & 78.02 $\pm$ 0.001         & 77.83 $\pm$ 0.007 & 76.16 $\pm$ 0.005 & 69.24 $\pm$ 0.007\\
PixelatedBFly   & Row & 78.12 $\pm$ 0.004 & 78.04 $\pm$ 0.011         & 77.49 $\pm$ 0.003 & 74.09 $\pm$ 0.002 & 62.82 $\pm$ 0.009\\
DSB             & Row & 78.09 $\pm$ 0.004 & 77.95  $\pm$ 0.005        & 76.34 $\pm$ 0.003 & 73.09 $\pm$ 0.004 & 64.49 $\pm$ 0.009\\
\method         & Row & 78.54 $\pm$ 0.006 & 78.26 $\pm$ 0.008         & 77.85 $\pm$ 0.006 & 77.19 $\pm$ 0.003 & 70.12 $\pm$ 0.010\\
\bottomrule
\end{tabular}}
\end{table*}

\begin{table*}[ht!]
\centering
\caption{Top-1 accuracy of sparse training methods at varying sparsities. We bold results that are not significantly different (based on paired asymptotic McNemar tests ($\alpha = 0.05$)) from the best-performing method (marked with a *) in each column. We see an increase in the generalization performance of all the structured sparse networks on the ImageNet-1K dataset.}
\label{tab:res:imgnet}
\resizebox{0.9\textwidth}{!}{
\begin{tabular}{l l l c c c c c c}
\toprule
\textbf{Model} & \textbf{Method} & \textbf{Perm.} &\textbf{Struc.} &\textbf{60\%} & \textbf{70\%} & \textbf{80\%} & \textbf{90\%} & \textbf{95\%} \\
\midrule
\multirow{18}{*}{\textbf{ViT-B/16}} 
&          &\multicolumn{7}{c}{\textit{dense accuracy = 78.5}}\\\cline{2-9}
& RigL & -                  &no             &\textbf{79.75}  & \textbf{79.28}   & \textbf{78.71}  & \textbf{77.24}     & \textbf{71.50} \\
& SET & -                   & no            &78.15           & 78.01            & 77.78           & \textbf{77.01}     & \textbf{71.48} \\
 & CHT & -                  & no            &\textbf{79.78}  &\textbf{79.37}    & 79.06*          & 77.66*             & 71.68* \\
 & CHTs & -                 & no            & 79.88*         &79.38*            & \textbf{79.05}  & \textbf{77.54}     & \textbf{71.61} \\
 & Mest  & -                & no            &78.04           &77.76             & 77.39           & 76.45              & 69.67 \\
& SRigl & -                 &\textbf{yes}   &77.79           & 77.84            & 77.35           & 75.90              & 68.70 \\
& PixelatedBFly & -         &\textbf{yes}   &78.04           &77.90             & 77.31           &73.89               & 62.52 \\
& DSB & -                   &\textbf{yes}   &77.98           & 77.85            & 76.26           &72.89               & 64.17 \\
& \method & -               &\textbf{yes}   &78.29          & 77.94             & 77.62    & \textbf{76.91}            & 69.54\\
& SRigL & Random            &\textbf{yes}   &77.95          & 77.81             & 77.31    & 75.69                     & 68.74 \\
& PixelatedBFly & Random    &\textbf{yes}   &77.91          & 77.94             & 77.34    & 73.93                     & 62.45 \\
& DSB & Random              &\textbf{yes}   &78.06          & 77.84             & 76.27    & 72.84                     & 64.23 \\
& \method & Random          &\textbf{yes}   &78.21          & 77.92             & 77.67    & 76.93                     & 69.54\\
& SRigL & \MethodName            &\textbf{yes}   &78.04          & 78.02             & 77.83    & 76.16                     & 69.24 \\
& PixelatedBFly & \MethodName    &\textbf{yes}   &78.10          & \textbf{78.04}    & 77.49    & 74.09                     & 62.82 \\
& DSB & \MethodName              &\textbf{yes}   &78.11          & 77.95             & 76.34    & 73.09                     & 64.49 \\
& \method & \MethodName          &\textbf{yes}   &\textbf{78.53} & \textbf{78.26}    & 77.85    & \textbf{77.19}            & 70.12\\
\midrule
\multirow{14}{*}{\textbf{ViT-L/16}}
&                 & \multicolumn{7}{c}{\textit{dense accuracy = 82.2}} \\ \cline{2-9}
& RigL            & -        & no          & 81.85*         & 81.57*            & 81.70*            & 78.26*        & 72.11* \\
& SRigL           & -        & yes         & 79.87          & 78.94             & 77.54             & 75.46         & 66.68  \\
& PixelatedBFly   & -        & yes         & 79.13          & 79.06             & 79.33             & 75.12         & 66.59  \\
& DSB             & -        & yes         & 79.44          & 77.46             & 75.34             & 73.55         & 66.77  \\
& \method         & -        & yes         & \textbf{81.52}  & \textbf{81.46}   & \textbf{81.37}    & 77.74  & 69.59  \\
& SRigL           & Random   & yes         & 79.94           & 78.95            & 77.55             & 75.56           & 66.74  \\
& PixelatedBFly   & Random   & yes         & 79.14           & 79.19            & 79.34             & 75.29           & 66.70  \\
& DSB             & Random   & yes         & 79.42           & 77.51            & 75.39             & 73.71           & 66.74  \\
& \method         & Random   & yes         & 80.21           & 80.16   & 79.52    & 75.31           & 69.56  \\
& SRigL           & \MethodName   & yes         & 80.29           & 79.63            & 77.96             & 76.78           & 66.86  \\
& PixelatedBFly   & \MethodName   & yes         & 79.33           & 79.23            & \textbf{80.33}    & 75.67           & 66.79  \\
& DSB             & \MethodName   & yes         & 79.59           & 77.61            & 75.51             & 73.79           & 66.91  \\
& \method         & \MethodName   & yes         & \textbf{81.66}  & \textbf{81.59}   & \textbf{81.49}    & \textbf{77.96}  & 70.16  \\
\midrule

\multirow{14}{*}{\textbf{Mixer-S/16}}
&                 & \multicolumn{7}{c}{\textit{dense accuracy = 72.4}} \\ \cline{2-9}
& RigL            & -        & no    & 73.21*           & 73.23*        & 73.47*            & 73.01*            & 70.41*  \\
& SRigL           & -        & yes   & 71.89            & 72.05         & 71.71             & 70.21             & 66.87  \\
& PixelatedBFly   & -        & yes   & 71.95            & 71.91         & 71.45             & 69.17             & 67.85  \\
& DSB             & -        & yes   & 69.94            & 70.21         & 68.90             & 65.16             & 60.88  \\
& \method         & -        & yes   & \textbf{72.92}  & 72.95          & \textbf{73.05}    & \textbf{72.31}    & \textbf{68.89}  \\
& SRigL           & Random   & yes   & \textbf{71.84}  & 72.07          & 71.74             & 70.21             & 66.86  \\
& PixelatedBFly   & Random   & yes   & 71.91            & 71.94         & 71.49             & 69.17             & 67.91  \\
& DSB             & Random   & yes   & 69.93            & 70.23         & 68.81             & 65.16             & 60.81  \\
& \method         & Random   & yes   & \textbf{72.93}  & 72.41          & 72.32             & 71.91             & 68.01  \\
& SRigL           & \MethodName   & yes   & \textbf{72.56}  & \textbf{72.91} & \textbf{72.79}    & 71.24             & 67.16  \\
& PixelatedBFly   & \MethodName   & yes   & 72.14            & 72.09         & 71.86             & 69.47             & \textbf{68.71}  \\
& DSB             & \MethodName   & yes   & 70.22            & 70.39         & 69.12             & 65.16             & 61.08  \\
& \method         & \MethodName   & yes   & \textbf{73.09}  & \textbf{73.11}  & \textbf{73.21}  & \textbf{72.51}     & \textbf{69.19}  \\

\bottomrule
\end{tabular}}
\end{table*}

\begin{table*}[t]
\centering
\caption{Perplexity of GPT-2 Small across sparsity levels and downstream task performance at 90\% sparsity. Avg Gap (PPL) reports the average relative \emph{increase} in perplexity over the unstructured baseline (RigL); lower is better. Downstream Avg is the mean across five zero-shot tasks; higher is better. \MethodName consistently reduces the perplexity gap below both the structured baseline and the random-permutation control, with parallel improvements in downstream performance. Full per-task downstream results at all sparsity levels are in Appendix~\ref{apx:over}.}
\label{tab:gpt_s_results}
\definecolor{padstgray}{gray}{0.93}
\definecolor{deltagreen}{HTML}{1F8A3D}
\resizebox{\textwidth}{!}{%
\begin{tabular}{lccccccrrrrrr}
\toprule
\textbf{Method}
& \multicolumn{5}{c}{\textbf{Perplexity ($\downarrow$)}} 
& \textbf{Avg Gap}
& \multicolumn{6}{c}{\textbf{Downstream at 90\% Sparsity ($\uparrow$)}} \\
\cmidrule(lr){2-6}
\cmidrule(lr){8-13}
& \textbf{40\%} & \textbf{50\%} & \textbf{60\%} & \textbf{80\%} & \textbf{90\%} & \textbf{(\%)}
& \textbf{LAMB.} & \textbf{ARC-E} & \textbf{ARC-C} & \textbf{HellaS.} & \textbf{WinoG.} & \textbf{Avg} \\
\midrule
\textit{Dense} & \multicolumn{5}{c}{22.21} & --
& 45.99 & 43.81 & 19.03 & 28.92 & 51.62 & 37.87 \\
\midrule
RigL & 22.34 & 22.80 & 23.79 & 29.87 & 53.76 & --
& 39.27 & 40.16 & 17.31 & 25.46 & 49.84 & 34.41 \\
\midrule

SRigL                  & 22.74 & 23.19 & 25.09 & 31.08 & 62.55 & 5.87
& 38.11 & 38.26 & 16.08 & 23.93 & 48.53 & 32.98 \\
\quad + Random         & 22.70 & 23.19 & 25.21 & 31.11 & 62.42 & 5.91
& 29.82 & 31.29 & 12.49 & 20.34 & 44.23 & 27.63 \\
\rowcolor{padstgray}
\quad + \MethodName    & 22.41 & 22.94 & 24.59 & 30.40 & 60.96 & 3.89 \,\textcolor{deltagreen}{$\downarrow$\,1.98}
& 38.62 & 39.49 & 16.87 & 24.38 & 49.14 & 33.70 \\
\midrule

PixelatedBFly          & 22.50 & 23.25 & 25.98 & 34.89 & 66.44 & 10.46
& 32.18 & 36.37 & 14.72 & 22.14 & 46.17 & 30.32 \\
\quad + Random         & 22.54 & 23.22 & 26.09 & 34.84 & 66.46 & 10.53
& 29.58 & 31.03 & 11.65 & 19.53 & 41.57 & 26.67 \\
\rowcolor{padstgray}
\quad + \MethodName    & 22.41 & 23.01 & 25.69 & 34.71 & 66.06 & 9.66 \,\textcolor{deltagreen}{$\downarrow$\,0.80}
& 34.74 & 37.33 & 15.21 & 22.47 & 47.34 & 31.42 \\
\midrule

DynaDiag               & 22.60 & 22.74 & 24.67 & 30.46 & 56.33 & 2.27
& 37.86 & 38.67 & 16.22 & 24.21 & 48.71 & 33.13 \\
\quad + Random         & 22.61 & 23.19 & 25.12 & 31.69 & 57.61 & 4.35
& 34.23 & 35.10 & 13.50 & 21.58 & 43.69 & 29.62 \\
\rowcolor{padstgray}
\quad + \MethodName    & 22.44 & 22.69 & 24.51 & 30.26 & 55.49 & 1.50 \,\textcolor{deltagreen}{$\downarrow$\,0.77}
& 38.44 & 39.79 & 17.07 & 24.33 & 49.53 & 33.83 \\
\bottomrule
\end{tabular}%
}
\end{table*}

\begin{table}[ht!]
\centering
\caption{Perplexity of sparse training methods at varying levels of sparsity. We bold results that are not significantly different from the best-performing method (marked with a *) based on paired asymptotic McNemar tests ($\alpha = 0.05$). We see an improvement in the PPL (lower the better) score for all structured sparse training methods with permutaitons on the WikiText-103 dataset.}
\label{tab:gpt2}
\resizebox{0.9\textwidth}{!}{
\begin{tabular}{l l l c c c c c}
\toprule
\textbf{Model} & \textbf{Method} & \textbf{Perm.} &  \textbf{40\%} & \textbf{50\%} & \textbf{60\%} & \textbf{80\%} & \textbf{90\%}\\
\midrule
\multirow{12}{*}{\textbf{GPT2-S}}
&                 & \multicolumn{6}{c}{\textit{dense PPL = 22.21}} \\ \cline{2-8}
& RigL            & -        & 22.34*            & \textbf{22.80}   & 23.79*            & 29.87*            & 53.76* \\
& SRigL           & -        & 22.74            & 23.19             & 25.09             & 31.08             & 62.55 \\
& PixelatedBFly   & -        & \textbf{22.50}   & 23.25             & 25.98             & 34.89             & 66.44 \\
& \method         & -        & 22.60            & \textbf{22.74}    & \textbf{24.67}    & \textbf{30.46}    & 56.33 \\
& SRigL           & Random   & 22.70            & 23.19             & 25.21             & 31.11             & 62.42 \\
& PixelatedBFly   & Random   & 22.54            & 23.22             & 26.09             & 34.84             & 66.46 \\
& \method         & Random   & 22.61            & 23.19             & 25.12             & 31.69             & 57.61 \\
& SRigL           & \MethodName   & \textbf{22.41}  & \textbf{22.94}     & \textbf{24.59}    & \textbf{30.40}    & 60.96 \\
& PixelatedBFly   & \MethodName   & \textbf{22.41}  & 23.01              & 25.69             & 34.71             & 66.06 \\
& \method         & \MethodName   & \textbf{22.44}  & 22.69*             & \textbf{24.51}    & \textbf{30.26}    & \textbf{55.49} \\
\midrule
\multirow{12}{*}{\textbf{GPT2-M}}
&                 & \multicolumn{6}{c}{\textit{dense PPL = 20.18}} \\ \cline{2-8}
& RigL            & -        & 20.45*           & 21.60*            & 23.49*            & 28.87*            & 51.76* \\
& SRigL           & -        & 21.14            & 22.59             & 26.09             & 32.16             & 55.66 \\
& PixelatedBFly   & -        & \textbf{20.86}  & 22.49              & 25.45             & 34.24             & 56.09 \\
& \method         & -        & \textbf{20.69}  & \textbf{22.14}     & \textbf{24.98}  & \textbf{29.65}      & 54.87 \\
& SRigL           & Random   & 21.19           & 22.55              & 26.01             & 32.19             & 55.69 \\
& PixelatedBFly   & Random   & 20.90            & 22.51             & 25.44             & 34.22             & 56.01 \\
& \method         & Random   & 21.65            & 22.67             & 25.17             & 30.39             & 54.81 \\
& SRigL           & \MethodName   & \textbf{20.57}  & \textbf{22.20}     & \textbf{25.04}    & \textbf{29.89}    & 55.13 \\
& PixelatedBFly   & \MethodName   & \textbf{20.69}  & \textbf{22.23}     & 25.31             & 33.19             & 55.71 \\
& \method         & \MethodName   & \textbf{20.55}  & \textbf{21.91}     & \textbf{24.71}    & \textbf{29.21}    & 54.26 \\

\bottomrule
\end{tabular}}
\end{table}

\subsection{Details of Layers Sparsified}
\label{sec:apx:layerDetails}
In our ViT-B/16 experiments, we applied sparsity to the initial patch projection layer, the MLP layers, and the output projections of the multi-head attention (MHA) modules. For the GPT models, we sparsified all attention and MLP layers. 


\section{Impact Statement}
\label{sec:imp}
Our work introduces a method to learn permutations for improving the performance of strucutred sparse neural networks. Structured sparse networks are useful because they can enable training larger models with fixed computational resources,  supporting the adoption of deep learning as model sizes continue to grow. 

While we anticipate primarily positive impacts through improved model performance and accessibility, we acknowledge that any advancement in machine learning capabilities warrants careful consideration. We recommend that any fugure research that builds on top of our work should investigate potential impacts on factors such as bias and fairness.





\end{document}